\documentclass[10pt,twocolumn,letterpaper]{article}

\usepackage{cvpr}
\usepackage{times}
\usepackage{epsfig}
\usepackage{graphicx}
\usepackage{subcaption}
\usepackage{multirow}
\usepackage{amsmath}
\usepackage{amssymb}
\usepackage{booktabs}
\usepackage{pifont}
\usepackage{textcomp}
\usepackage[dvipsnames]{xcolor}
\usepackage{multicol}
\usepackage[title]{appendix}
\usepackage[pagebackref=true,breaklinks=true,letterpaper=true,colorlinks,bookmarks=false]{hyperref}

\cvprfinalcopy

\def\cvprPaperID{0000}

\newcommand{\textapprox}{\raisebox{0.3ex}{\texttildelow}}
\ifcvprfinal\fi

\begin{document}
\newcommand{\modelname}{\textsc{Elastic}}

\title{ELASTIC: Improving CNNs with Dynamic Scaling Policies}

\author{
Huiyu Wang$^{1}$\thanks{Work done while an intern at AI2.}~~~Aniruddha Kembhavi$^{2}$~~~Ali Farhadi$^{2,3,4}$~~~Alan Yuille$^{1}$~~~Mohammad Rastegari$^{2,4}$\\
$^{1}$Johns Hopkins University~~~~$^{2}$PRIOR @ Allen Institute for AI\\$^{3}$University of Washington~~~~$^{4}$Xnor.ai\\
\small \url{ https://prior.allenai.org/projects/elastic} \\
}
\maketitle

\begin{abstract}
Scale variation has been a challenge from traditional to modern approaches in computer vision. Most solutions to scale issues have a similar theme: a set of intuitive and manually designed policies that are generic and fixed (e.g. SIFT or feature pyramid). We argue that the scaling policy should be learned from data. In this paper, we introduce \modelname, a simple, efficient and yet very effective approach to learn a dynamic scale policy from data. We formulate the scaling policy as a non-linear function inside the network's structure that (a) is learned from data, (b) is instance specific, (c) does not add extra computation, and (d) can be applied on any network architecture.  We applied \modelname~ to several state-of-the-art network architectures and showed consistent improvement without extra (sometimes even lower) computation on ImageNet classification, MSCOCO multi-label classification, and PASCAL VOC semantic segmentation. Our results show major improvement for images with scale challenges. Our code is available here: \url{https://github.com/allenai/elastic} 
\end{abstract}

\section{Introduction}

Scale variation has been one of the main challenges in computer vision. There is a rich literature on different approaches to encoding scale variations in computer vision algorithms ~\cite{lindeberg1994scale}. In feature engineering, there have been manually prescribed solutions that offer scale robustness. For example, the idea of searching for scale first and then extracting features based on a known scale used in SIFT or the idea of using feature pyramids are examples of these prescribed solutions. Some of these ideas have also been migrated to feature learning using deep learning in modern recognition solutions. 
\begin{figure}[t]
\begin{center}
   \includegraphics[width=0.98\linewidth]{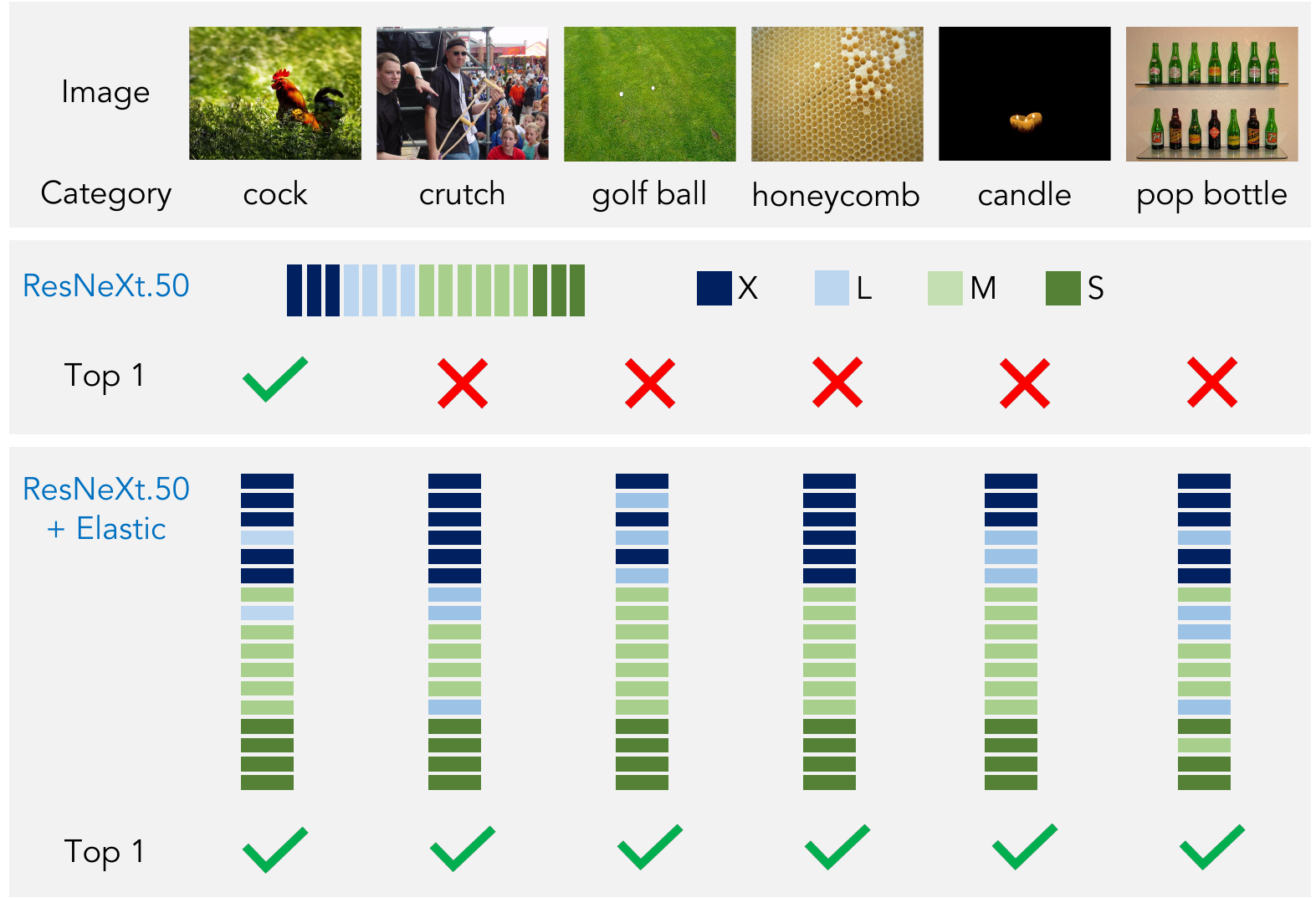}
\end{center}
\vspace{-0.4cm}
\caption{\textbf{Dynamic scale policy.} Scaling policies in CNNs are typically integrated into the network architecture manually in a pyramidal fashion. The color bar in this figure (second row) shows the scales at different blocks of the ResNext50 architecture. The early layers receive {\color{Blue}e\textbf{X}tra-large} resolutions and in the following layers resolutions decrease as {\color{ProcessBlue}\textbf{L}arge}, {\color{LimeGreen}\textbf{M}edium}, and {\color{ForestGreen}\textbf{S}mall}. We argue that scaling policies in CNNs should be instance-specific. Our Elastic model (the third row) allows different scaling policies for different input images and it learns from the training data how to pick the best policy. For scale challenging images e.g. images with lots of small(or diverse scale) objects, it is crucial that network can adapt its scale policy based on the input. As it can be seen in this figure, Elastic gives a better prediction for these scale challenging images. (See section~\ref{sec:policyanal} for more details)}
\label{fig:teaser}
\vspace{-0.4cm}
\end{figure}

The majority of the solutions in old-school and even modern approaches to encode scale are manually designed and fixed solutions. For example, most state-of-the-art image classification networks \cite{krizhevsky2012imagenet, simonyan2014very, he2016deep, huang2017densely, yu2018deep, zoph2016neural} use the feature pyramid policy where a network looks at the larger resolution first and then goes to smaller ones as it proceeds through the layers. Despite the fact that this common practice seems to be a natural and intuitive choice, we argue that this scale policy is not necessarily the best one for all possible scale variations in images. We claim that an ideal scale policy should  (1) be learned from the data; (2) be instance specific; (3) not add extra computational burden; and (4) be applicable to any network architecture. 

For example, instead of looking at the scales according to the feature pyramid policy if we process the images in Figure~\ref{fig:teaser} based on a learned and instance specific policy we see an improved performance. In images with scale challenges like the golf ball image in Figure~\ref{fig:teaser} the learned scale policy might differ dramatically from a pyramid policy, resulting in correct classification of that instance. The learned policy for this instance starts from looking at the image from a large scale (dark blue color), and then goes immediately to a smaller scale, and then goes back to a large scale followed by a small scale and so on. 

In this paper, we introduce \modelname, an approach to learn instance-specific and not-necessarily-pyramidal scale policies with no extra(or lower) computational cost. Our solution is simple, efficient, and very effective on a wide range of network architectures for image classification and segmentation. Our Elastic model can be applied on any CNN architectures simply by adding downsamplings and upsamplings in parallel branches at each layer and let the network learn from data a scaling policy in which inputs being processed at different resolutions in each layer. We named our model \modelname~ because each layer in the network is flexible in terms of choosing the best scale by a soft policy.    

Our experimental evaluations show improvements in image classification on ImageNet\cite{russakovsky2015imagenet}, multi-label classification on MSCOCO\cite{lin2014microsoft}, and semantic segmentation on PASCAL VOC for ResNeXt\cite{xie2017aggregated}, SE-ResNeXt\cite{hu2017squeeze}, DenseNet\cite{huang2017densely}, and Deep Layer Aggregation (DLA)\cite{yu2018deep} architectures. Furthermore, our results show major improvements (about 4\%) on images with scale challenges (lots of small objects or large variation across scales within the same image) and lower improvements for images without scale challenges. Our qualitative analysis shows that images with similar scaling policies (over the layers of the network) are sharing similar complexity pattern in terms of scales of the objects appearing in the image.    
\section{Related Work}
\label{sec:relatedwork}
\begin{figure*}[t!]
\begin{center}
   \includegraphics[width=0.98\textwidth]{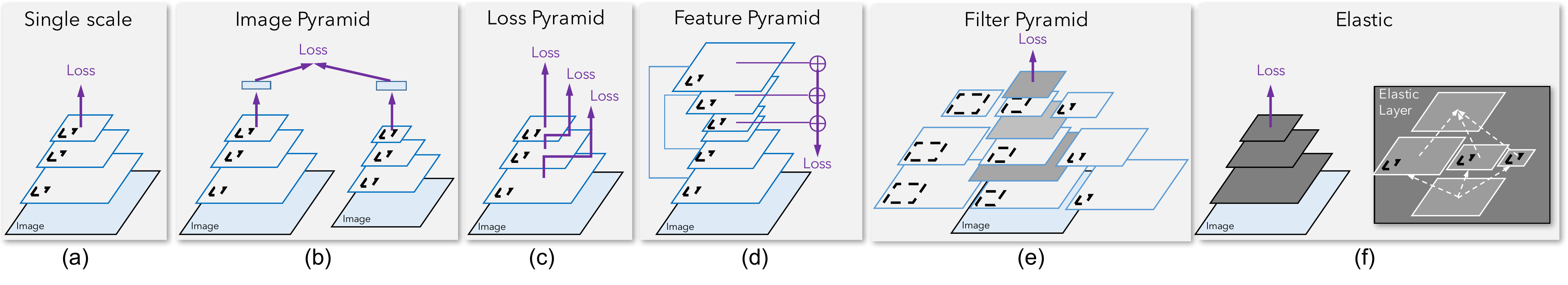}
\end{center}
\vspace{-0.5cm}
\caption{\textbf{Multi-scaling model structures.} This figure illustrates different approaches to multi-scaling in CNN models and our Elastic model. The solid-line rectangles show the input size and the dashed-line rectangles shows the filter size.}
\label{fig:multiscale}
\vspace{-0.2cm}
\end{figure*}

The idea behind Elastic is conceptually simple and there are several approaches in the literature using similar concepts. Therefore, we study all the categories of related CNN models and clarify the differences and similarities to our model. There are several approaches to fusing information at different visual resolutions. The majority of them are classified into four categories (depicted in Figure~\ref{fig:multiscale}(b-e)).

\textbf{Image pyramid}: An input image is passed through a model multiple times at different resolutions and predictions are made independently at all levels. The final output is computed as an ensemble of outputs from all resolutions. This approach has been a common practice in \cite{dalal2005histograms, felzenszwalb2010object, sermanet2013overfeat}.

\textbf{Loss pyramid}: This method enforces multiple loss functions at different resolutions. \cite{szegedy2015going} uses this approach to improve the utilization of computing resources inside the network. SSD \cite{liu2016ssd} and MS-CNN \cite{cai2016unified} also use losses at multiple layers of the feature hierarchy.

\textbf{Filter pyramid}: Each layer is divided into multiple branches with different filter sizes (typically referred to as the split-transform-merge architecture). The variation in filter sizes results in capturing different scales but with additional parameters and operations. The inception family of networks \cite{szegedy2015going,szegedy2016rethinking,szegedy2017inception} use this approach. To further reduce the complexity of the filter pyramid \cite{mehta2018espnet,yu2015multi,yu2017dilated} use dilated convolutions to cover a larger receptive field with the same number of FLOPs. In addition, \cite{chen2018learning} used 2 CNNs to deal with high and low frequencies, and \cite{zhou2017adaptive} proposed to adaptively choose from 2 CNNs with different capacity.

\textbf{Feature pyramid}: This is the most common approach to incorporate multiple scales in a CNN architecture. Features from different resolutions are fused in a network by either concatenation or summation. Fully convolutional networks \cite{long2015fully} add up the scores from multiple scales to compute the final class score. Hypercolumns \cite{hariharan2015hypercolumns} use earlier layers in the network to capture low-level information and describe a pixel in a vector. Several other approaches (HyperNet \cite{kong2016hypernet}, ParseNet \cite{liu2015parsenet}, and ION \cite{bell2016inside}) concatenate the outputs from multiple layers to compute the final output. Several recent methods including SharpMask \cite{pinheiro2016learning} and U-Net \cite{ronneberger2015u} for segmentation, Stacked Hourglass networks \cite{newell2016stacked} for keypoint estimation and Recombinator networks \cite{honari2016recombinator} for face detection, have used skip connections to incorporate low-level feature maps on multiple resolutions and semantic levels. \cite{huang2017condensenet} extends DenseNet\cite{huang2017densely} to fuse features across different resolution blocks. Feature pyramid networks (FPNs) \cite{lin2017feature} are designed to normalize resolution and equalize semantics across the levels of a pyramidal feature resolution hierarchy through top-down and lateral connections. Likewise, DLA~\cite{yu2018deep} proposes an iterative and hierarchical deep aggregation that fuses features from different resolutions.

Elastic resembles models from the Filter pyramid family as well as the Feature pyramid family, in that it introduces parallel branches of computation (a la Filter pyramid) and also fuses information from different scales (a la Feature pyramid). The major difference to the feature pyramid models is that in Elastic every layer in the network considers information at multiple scales uniquely whereas in feature pyramid the information for higher or lower resolution is injected from the other layers. Elastic provides an exponential number of scaling paths across the layers and yet keeps the computational complexity the same (or even lower) as the base model. The major difference to the filter pyramid is that the number of FLOPs to cover a higher receptive field in Elastic is proportionally lower, due to the downsampling whereas in the filter pyramid the FLOPs is higher or the same as the original convolution. 
\section{Model}
\label{sec:model}

In this section, we elaborate the structure of our proposed Elastic and illustrate standard CNN architectures being augmented with our Elastic. We also contrast our model with other multi-scale approaches. 

\subsection{Scale policy in CNN blocks}
Formally, a layer in a CNN can be expressed as 
\vspace{-0.2cm}
\begin{equation} 
\mathcal{F}(x) = \sigma \Big(\sum_{i=1}^{q} \mathcal{T}_i(x)\Big)
\end{equation} 
where $q$ is the number of branches to be aggregated, $\mathcal{T}_{i}(x)$ can be an arbitrary function (normally it is a combination of convolution, batch normalization and activation), and $\sigma$ are nonlinearities.
A few $\mathcal{F}(x)$ are stacked into a stage to process information in one spatial resolution. Stages with decreasing spatial resolutions are stacked to integrate a pyramid scale policy in the network architecture. A network example of 3 stages with 2 layers in each stage is
\begin{equation} 
\label{eq:recblockdownsample}
\mathcal{N} = \mathcal{F}_{32} \circ \mathcal{F}_{31} \circ \mathcal{D}_{r_2} \circ \mathcal{F}_{22} \circ \mathcal{F}_{21} \circ \mathcal{D}_{r_1} \circ \mathcal{F}_{12} \circ \mathcal{F}_{11}
\end{equation} 
 where $\mathcal{D}_{r_i}$ indicates the resolution decrease by ratio $r_i>1$ after a few layers. $\mathcal{D}_{r_i}$ can be simply implemented by increasing the stride in the convolution right after. For example, ResNeXt\cite{xie2017aggregated} stacks bottleneck layers in each resolution and use convolution with stride 2 to reduce spatial resolution.
This leads to a fixed scaling policy that enforces a linear relationship between number of layers and the effective receptive field of those layers.
Parameters of $\mathcal{T}_{i}(x)$ and the elements in input tenors $x$ are all of the tangible ingredients in a CNN that define computational capacity of the model. Under a fixed computational capacity measured by FLOPs, to improve the accuracy of such a model, we can either increase number of parameters in $\mathcal{T}_{i}(x)$ and decrease the resolution of $x$  or increase the resolution of $x$ and decrease number of parameters in $\mathcal{T}_{i}(x)$. By adjusting the input resolutions at each layer and number of parameters, we can define a scaling policy across the network. We argue that finding the optimal scaling policy (a trade-off between the resolution and number of parameters in each layer) is not trivial. There are several model designs toward increasing the accuracy and manually injecting variations of feature pyramid but most of them are at the cost of higher FLOPs and more parameters in the network. In the next section, we explain our solution that can learn an optimal scaling policy and maintain or reduce number of parameters and FLOPs while improving the accuracy.     
\subsection{The ELASTIC structure}

In order to learn image features at different scales, we propose to add down-samplings and up-samplings in parallel branches at each layer and let the network make decision on adjusting its process toward various resolutions at each layer. Networks can learn this policy from training data. We add down-samplings and up-samplings in parallel branches at each layer and divide all the parameters across these branches as follows:  

\begin{equation} 
\mathcal{F}(x) = \sigma \Big(\sum_{i=1}^{q} \mathcal{U}_{r_i}(\mathcal{T}_{i}(\mathcal{D}_{r_i}(x)))\Big)
\end{equation} 
\begin{equation} 
\mathcal{N} = \mathcal{F}_{32} \circ \mathcal{F}_{31}\circ \mathcal{F}_{22} \circ \mathcal{F}_{21} \circ \mathcal{F}_{12} \circ \mathcal{F}_{11}
\end{equation}
where $\mathcal{D}_{r_i}(x)$ and $\mathcal{U}_{r_i}(x)$ are respectively downsampling and upsampling functions which change spatial resolutions of features in a layer. Unlike in equation \ref{eq:recblockdownsample}, a few $\mathcal{F}$ are applied sequentially without downsampling the main stream, and $\mathcal{N}(x)$ has exactly the same resolution as original x.

Note that the learned scaling policy in this formulation will be instance-specific i.e. for different image instances, the network may activate branches in different resolutions at each layer. In section \ref{sec:experiments} we show that this instance-specific scaling policy improves prediction on images with scale challenges e.g. images consist of lots of small objects or highly diverse object sizes.    

Conceptually, we propose a new structure where information is always kept at a high spatial resolution, and each layer or branch processes information at a lower or equal resolution. In this way we decouple feature processing resolution ($\mathcal{T}_{i}$ processes information at different resolutions) from feature storage resolution (the main stream resolution of the network). 
This encourages the model to process different scales separately at different branches in a layer and thus capture cross-scale information. More interestingly, since we apply Elastic to almost all blocks, the dynamic combination of multiple scaling options at each layer leads to exponentially many different scaling paths. They interpolate between the largest and the smallest possible scale and collectively capture various scales. In fact, this intuition is aligned with our experiments, where we have observed different categories of images adopt different scaling paths (see section \ref{sec:policyanal}). For example, categories with clean and uniform background images mostly choose the low-resolution paths across the network and categories with complex and cluttered objects and background mostly choose the high-resolution paths across the network. 

The computational cost of our Elastic model is equal to or lower than the base model, because at each layer the maximum resolution is the original resolution of the input tensor. Low resolution branches reduce the computation and give us extra room for adding more layers to match the computation of the original model. 

This simple add-on of downsamplings and upsamplings (Elastic) can be applied to any CNN layers $\mathcal{T}_{i}(x)$ in any architecture to improve accuracy of a model. Our applications are introduced in the next section.

\begin{figure*}[t]
\begin{center}
   \includegraphics[width=0.98\textwidth]{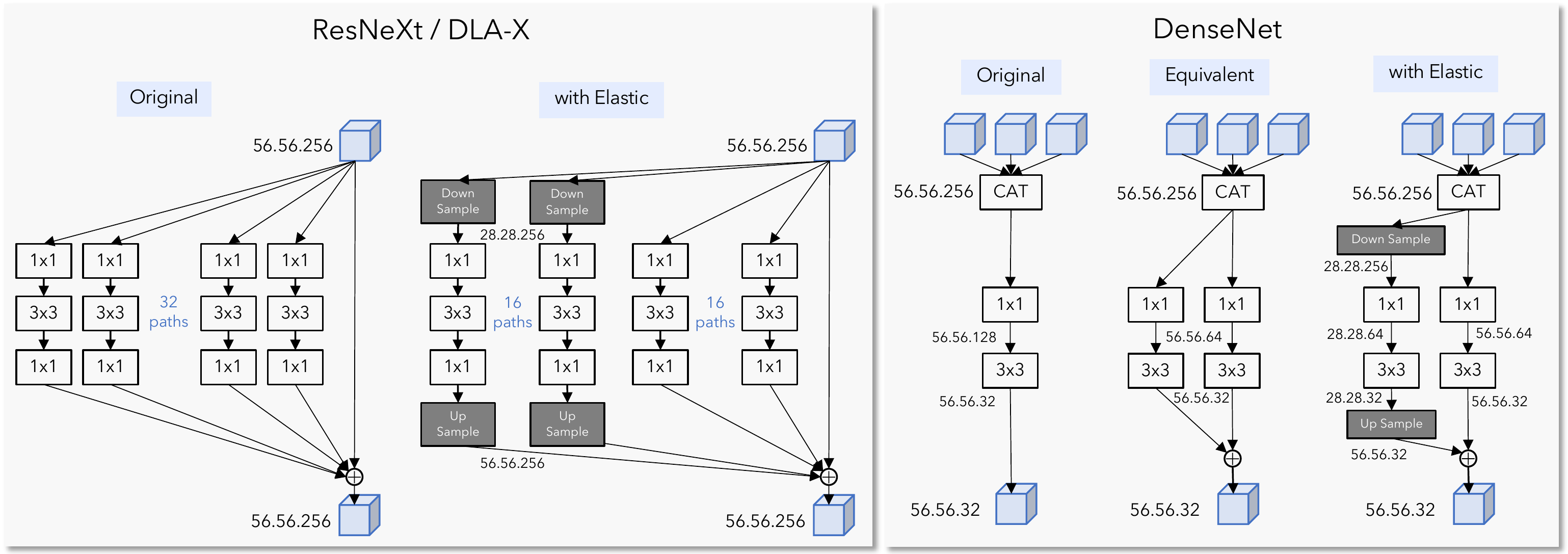}
\end{center}
\vspace{-0.5cm}
\caption{\textbf{Left:} ResNeXt bottleneck vs. Elastic bottleneck. \textbf{Right:} DenseNet block vs. its equivalent form vs. Elastic block. Elastic blocks spend half of the paths processing downsampled inputs in a low resolution, then the processed features are upsampled and added back to features with the original resolution. Elastic blocks have the same number of parameters and less FLOPs than original blocks}
\label{fig:network_extensions}
\vspace{-0.2cm}
\end{figure*}

\subsection{Augmenting models with Elastic}
Now, we show how to apply Elastic on different network architecture. To showcase the power of Elastic, we apply Elastic on some state-of-the-art network architectures: ResNeXt\cite{xie2017aggregated}, Deep Layer Aggregation (DLA)\cite{yu2018deep}, and DenseNet\cite{huang2017densely}. A natural way of applying Elastic on current classification models is to augment bottleneck layers with multiple branches. This makes our modification on ResNeXt and DLA almost identical. At each layer we apply downsampling and bilinear upsampling to a portion of branches, as shown in Figure \ref{fig:network_extensions}-left. In DenseNet we compile an equivalent version by parallelizing a single branch into two branches and then apply downsampling and upsampling on some of the branches, as shown in Figure \ref{fig:network_extensions}-right. Note that applying Elastic reduces FLOPs in each layer. To match the original FLOPs we increase number of layers in the network while dividing similar number of FLOPs across resolutions.

\vspace{-0.4cm}\paragraph{Relation to other multi-scaling approaches}
As discussed in section \ref{sec:relatedwork}, most of current multi-scaling approaches can be categorized into four different categories (1) \emph{image pyramid}, (2) \emph{loss pyramid} (3) \emph{filter pyramid}, and (4) \emph{feature pyramid}. Figure \ref{fig:multiscale}(b-e) demonstrates the structure of these categories. 
All of these models can improve the accuracy usually under a higher computational budget. Elastic (Figure ~\ref{fig:multiscale}) guarantees no extra computational cost while achieving better accuracy. Filter pyramid is the most similar model to Elastic. The major difference to the filter pyramid is that the number of FLOPs to cover a higher receptive field in Elastic is proportionally lower due to the downsampling whereas in the filter pyramid the FLOPs is higher or the same as the original convolution depending of filter size or dilation parameters. Table \ref{tab:scalecompare} compares the FLOPs and number of parameters between Elastic and feature/filter pyramid for a single convolutional operation. Note that the FLOPs and parameters in Elastic is always (under any branching $q$ and scaling ratio $r$) lower or equal to the original model whereas in filter/feature pyramid this is higher or equal. Feature pyramid methods are usually applied on top of an existing classification model, by concatenating features from different resolutions. It is capable of merging features from different scales in the backbone model and shows improvements on various tasks, but it does not intrinsically change the scaling policy. Our Elastic structure can be viewed as a feature pyramid inside a layer, which is able to model different scaling policies. Spatial pyramid pooling or Atrous(dilated) spatial pyramid shares the same limitation as feature pyramid methods.

\begin{table}[t!]
    \centering
    \resizebox{\columnwidth}{!}{
    \begin{tabular}{l|cc}
        \toprule
        \textbf{Multi-Scaling Method} & \textbf{FLOPs} & \textbf{Parameters}\\
        \midrule        
        Single Scale & \footnotesize $n^2ck^2$ & \footnotesize $ck^2$\\
        Feature Pyramid (concat) & \footnotesize $n^2(qc)k^2$ & \footnotesize $(qc)k^2$\\  
        Feature Pyramid (add) & \footnotesize $n^2ck^2$ & \footnotesize $ck^2$\\
        Filter Pyramid (standard) & \footnotesize $\sum_{i=1}^{q}\frac{n^2c(kr_i)^2}{b_i}$ & \footnotesize $\sum_{i=1}^{q}\frac{c(kr_i)^2}{b_i}$\\
        Filter Pyramid (dilated) & \footnotesize $n^2ck^2$ & \footnotesize $ck^2$\\
        Elastic & \footnotesize $\sum_{i=1}^{q}\frac{(\frac{n}{r_i})^2ck^2}{b_i}$ & \footnotesize $ck^2$\\
        \bottomrule
    \end{tabular}
    }
    \caption{\textbf{Computation in multi-scaling models.} This table compares the FLOPs and number of parameters between Elastic and feature/filter pyramid for a single convolutional operation, where the input tensor is $n\times n\times c$ and the filter size is $k\times k$. $q$ denotes the number of branches in the layer, where $\sum_{1}^{q}\frac{1}{b_i}=1$ and $b_i>1$ and $r_i>1$ denote the branching and scaling ratio respectively. Note that the FLOPs and parameters in Elastic is always (under any branching $q$ and scaling ratio $r$) lower than or equal to the original model whereas in feature/filter pyramid is higher or equal.}
    \label{tab:scalecompare}
    \vspace{-0.2cm}
\end{table}
\section{Experiments}
\label{sec:experiments}
In this section, we present experiments on applying Elastic to current strong classification models. We evaluate their performances on ImageNet classification, and we show consistent improvements over current models. Furthermore, in order to show the generality of our approach, we transfer our pre-trained Elastic models to multi-label image classification and semantic segmentation. We use ResNeXt \cite{xie2017aggregated}, DenseNet\cite{huang2017densely} and DLA \cite{yu2018deep} as our base models to be augmented with Elastic. 
\vspace{-0.4cm}\paragraph{Implementation details.} 
We use the official PyTorch ImageNet codebase with random crop augmentation but without color or lighting augmentation, and we report standard 224$\times$224 single crop error on the validation set. We train our model with 8 workers (GPUs) and 32 samples per worker. Following DLA \cite{yu2018deep}, all models are trained for 120 epochs with learning rate 0.1 and divided by 10 at epoch 30, 60, 90. We initialize our models using normal He initialization \cite{he2015delving}. Stride-2 average poolings are adopted as our downsamplings unless otherwise notified since most of our downsamplings are 2$\times$ downsamplings, in which case bilinear downsampling is equivalent to average pooling. Also, Elastic add-on is applied to all blocks except stride-2 ones or high-level blocks operating at resolution 7.
\subsection{ImageNet classification}
We evaluate Elastic on ImageNet\cite{russakovsky2015imagenet} 1000 way classification task (ILSVRC2012). The ILSVRC 2012 dataset contains 1.2 million training images and 50 thousand validation images. In this experiment, we show that our Elastic add-on consistently improves the accuracy of the state-of-the-art models without introducing extra computation or parameters. Table~\ref{tab:imagenet} compares the top-1 and top-5 error rates of all of the base models with the Elastic augmentation (indicated by '\textbf{+Elastic}')  and shows the number of parameters and FLOPs used for a single inference. Besides DenseNet, ResNeXt, DLA, SE-ResNeXt50+Elastic is also reported. In all the tables "*" denotes our implementation of the model. It shows that our improvement is almost orthogonal to the channel calibration proposed in \cite{hu2017squeeze}. In addition, we include ResNeXt50x2+Elastic to show that our improvement does not come from more depth added to ResNeXt101. In Figure 4 we project the numbers in the Table~\ref{tab:imagenet} into two plots: accuracy vs. number of parameters (Figure 4-left) and accuracy vs. FLOPs (Figure 4-right). This plot shows that our Elastic model can reach to a higher accuracy without any extra (or with lower) computational cost.    
\begin{figure}[t!]
    \centering
    \begin{subfigure}[b]{0.23\textwidth}
        \includegraphics[width=4cm]{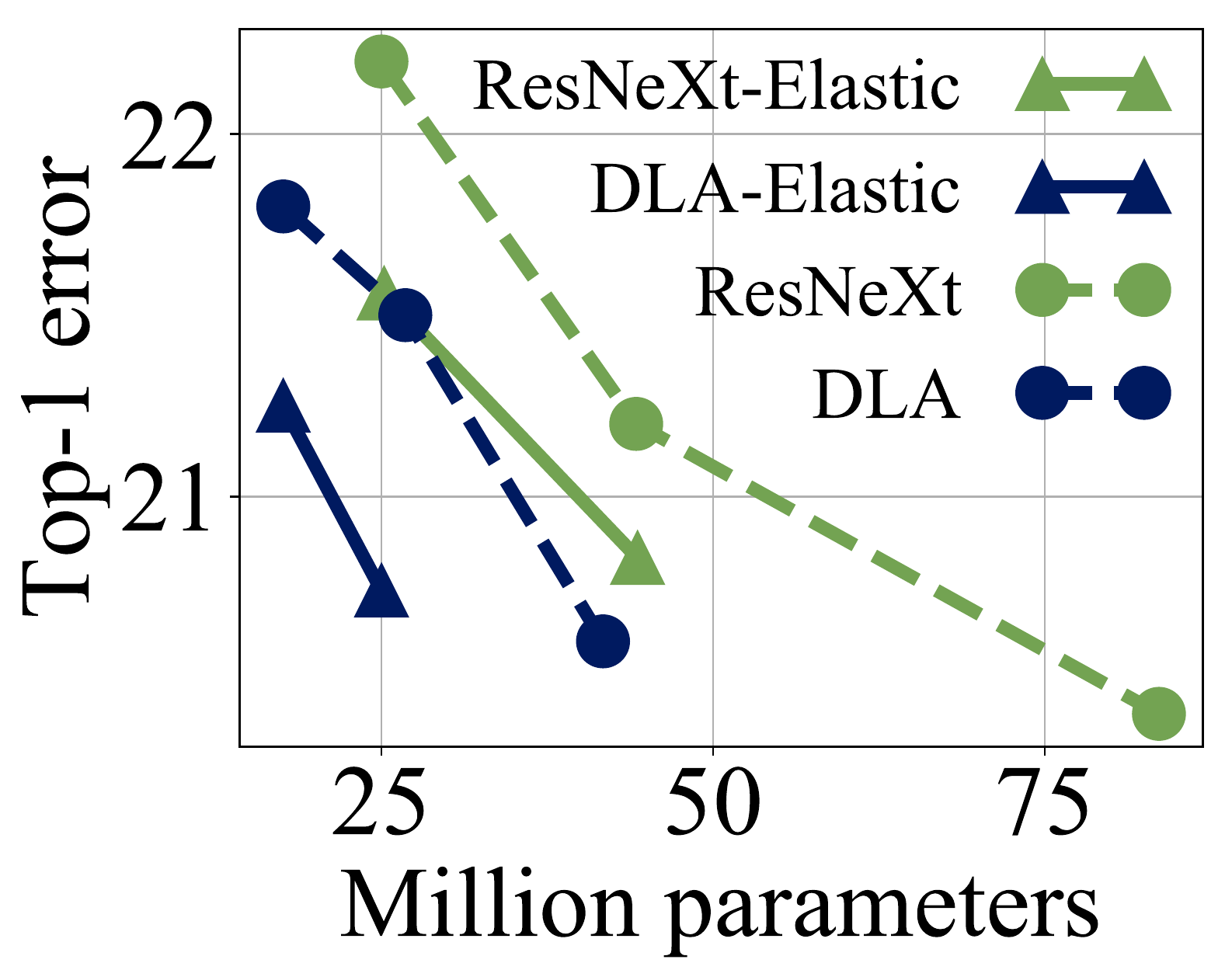}
        \label{fig:imgnetparam}
    \end{subfigure}
    \begin{subfigure}[b]{0.23\textwidth}
        \includegraphics[width=4cm]{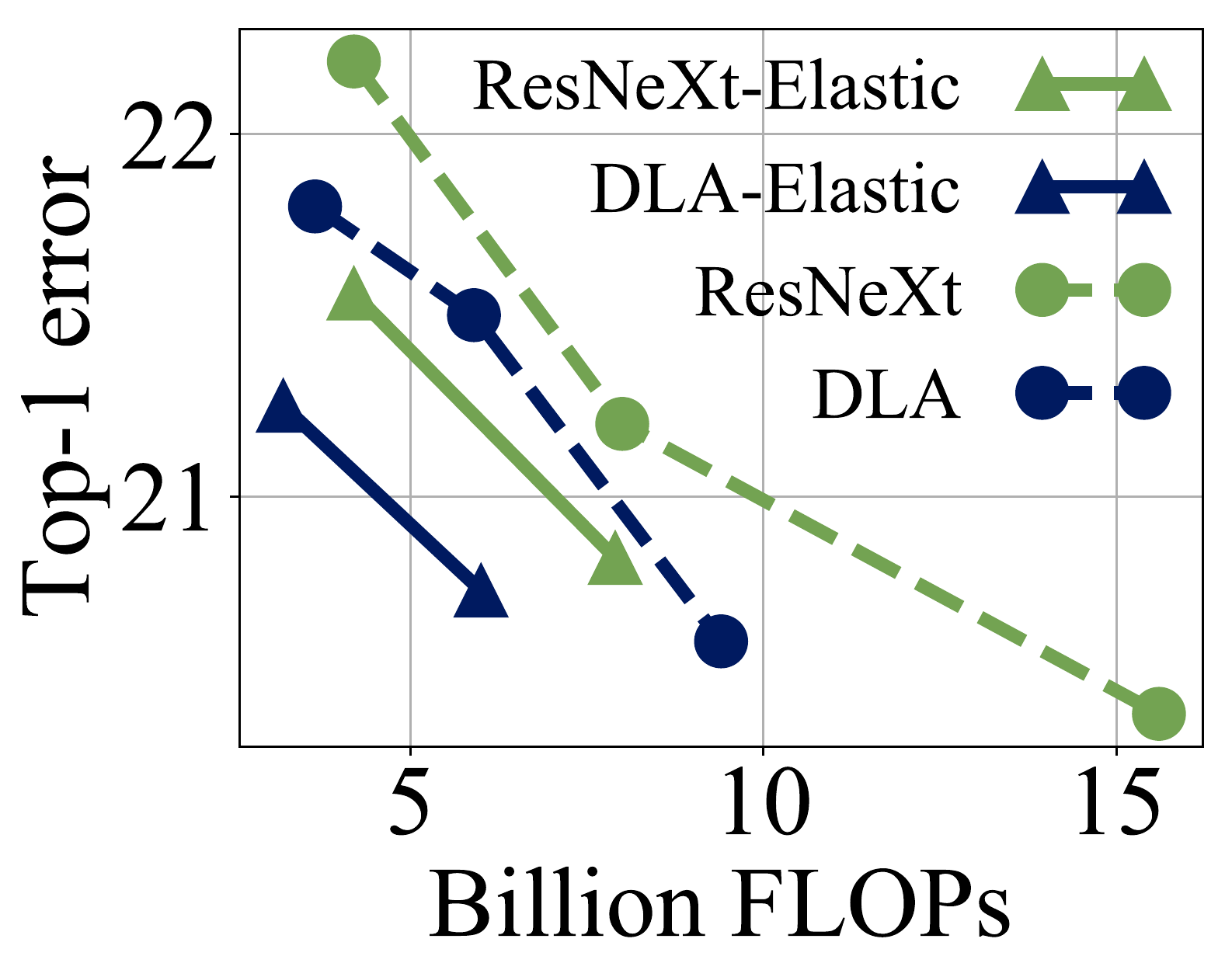}
        \label{fig:imgnetflop}
    \end{subfigure}
    \vspace{-0.5cm}
    \caption{\small\textbf{Imagenet Accuracy vs. FLOPS and Parameters} This figure shows our Elastic model can achieve a lower error without any extra (or with lower) computational cost.}
    \label{fig:plot}
    \vspace{-0.2cm}
\end{figure}
\begin{table}
\centering
\resizebox{0.9\columnwidth}{!}{
\setlength\tabcolsep{3pt}
\begin{tabular}{l|cc|cc}
\toprule
Model & \# Params & FLOPs & Top-1 & Top-5 \\
\midrule
DenseNet201\textsuperscript{*} & 20.0M & 4.4B & 22.25 & 6.26 \\
DenseNet201+Elastic & 19.5M & 4.3B & \textbf{22.07} & \textbf{6.00} \\
\midrule
ResNeXt50 & 25.0M & 4.2B & 22.2 & - \\
ResNeXt50\textsuperscript{*} & 25.0M & 4.2B & 22.23 & 6.25 \\
ResNeXt50+Elastic & 25.2M & 4.2B & \textbf{21.56} & \textbf{5.83} \\
\midrule
SE-ResNeXt50\textsuperscript{*} & 27.6M & 4.2B & 21.87 & 5.93 \\
SE-ResNeXt50+Elastic & 27.8M & 4.2B & \textbf{21.38} & \textbf{5.86} \\
\midrule
ResNeXt101 & 44.2M & 8.0B & 21.2 & 5.6 \\
ResNeXt101\textsuperscript{*} & 44.2M & 8.0B & 21.18 & 5.83 \\
ResNeXt101+Elastic & 44.3M & 7.9B & \textbf{20.83} & \textbf{5.41} \\
ResNeXt50x2+Elastic & 45.6M & 7.9B & 20.86 & 5.52 \\
\midrule
DLA-X60 & 17.6M & 3.6B & 21.8 & - \\ 
DLA-X60\textsuperscript{*} & 17.6M & 3.6B & 21.92 & 6.03 \\
DLA-X60+Elastic & 17.6M & 3.2B & \textbf{21.25} & \textbf{5.71} \\
\midrule
DLA-X102 & 26.8M & 6.0B & 21.5 & - \\
DLA-X102+Elastic & 25.0M & 6.0B & \textbf{20.71} & \textbf{5.38} \\
\bottomrule
\end{tabular}
}
\vspace{-0.2cm}
\caption{\small\textbf{State-of-the-art model comparisons on ImageNet validation set.} Base models (DenseNet, ResNeXt, and DLA) are augmented by Elastic (indicated by '+Elastic'). * indicates our implementation of these models. Note that augmenting with Elastic always improves accuracy across the board.}
\label{tab:imagenet}
\vspace{-0.2cm}
\end{table}
\subsubsection{Scale policy analysis}
\label{sec:policyanal}
\begin{figure}[t]
\begin{center}
   \includegraphics[width=0.45\textwidth]{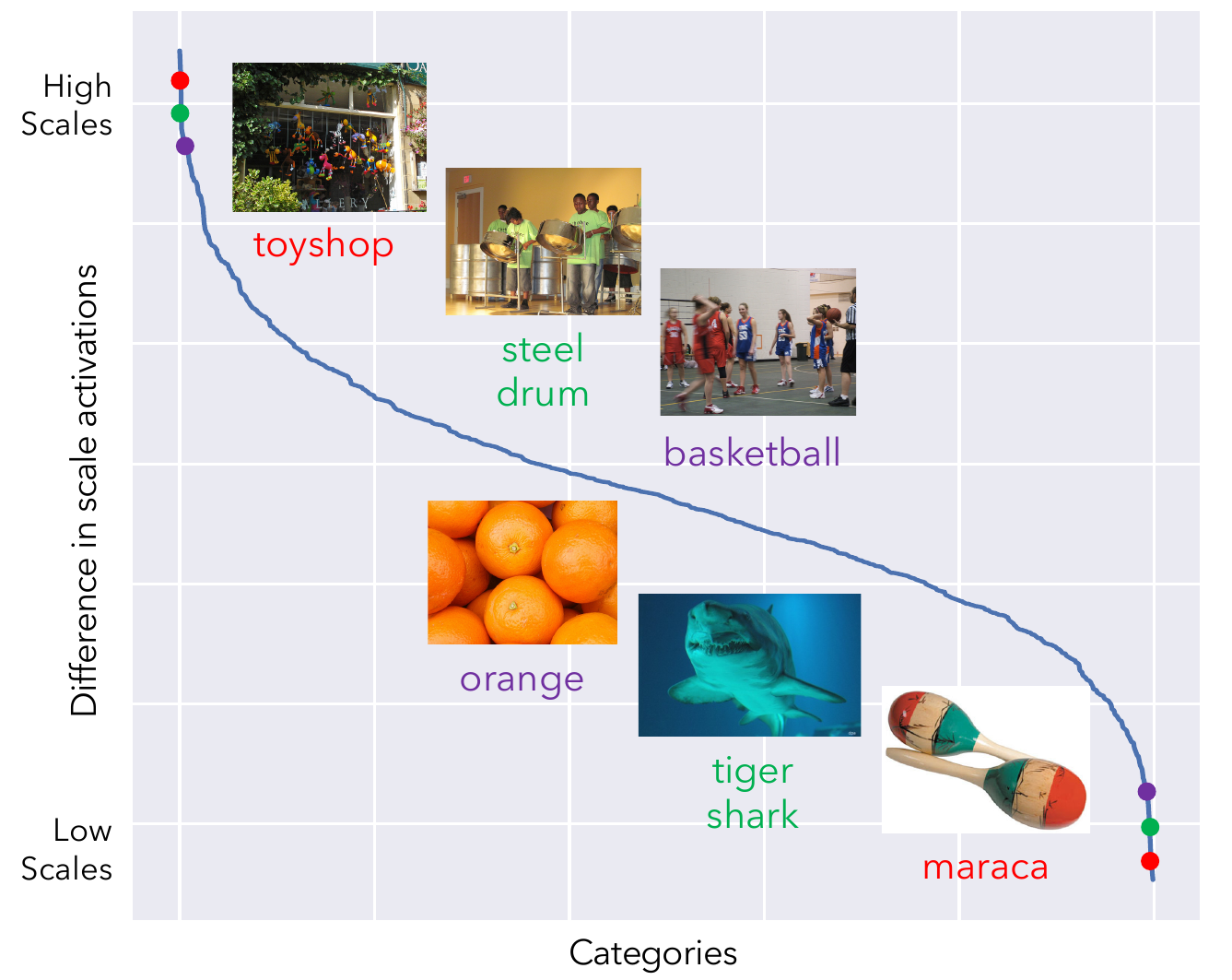}
\end{center}
\vspace{-0.5cm}\caption{\small\textbf{Scale policy for complex vs. simple image categories}. This figure shows the overall block scale policy score on the entire ImageNet categories. It shows that categories with complex image patterns mostly go through the high-resolution branches in the network and categories with simpler image pattern go through the low-resolution branches.}
\label{fig:category_activations}
\vspace{-0.2cm}
\end{figure}

To analyze the learned scale policy of our Elastic model, we define a simple score that shows at each block what was the resolution level (high or low) that the input tensor was processed. We formally define this scale policy score at each block by differences of mean activations in high-resolution and low-resolution branches.
\begin{equation}
\footnotesize
S = \frac{1}{4HWC} \sum_{h=1}^{2H} \sum_{w=1}^{2W} \sum_{c=1}^{C} x_{hwc}^{high}-\frac{1}{HWC} \sum_{h=1}^{H} \sum_{w=1}^{W} \sum_{c=1}^{C} x_{hwc}^{low}
\end{equation}
where $H$, $W$, $C$ are the height, width and number of channels in low resolution branches. $x^{high}$ and $x^{low}$ are the activations after $3\times3$ convolutions, fixed batch normalizations, and ReLU in high-resolution and low-resolution branches respectively. Figure ~\ref{fig:category_activations} shows all of the categories in ImageNet validation sorted by the mean scale policy score $S$ (average over all layers for all images within a category). As it can be seen, categories with more complex images appear to have a larger $S$ i.e. they mostly go through high-resolution branches in each block and images with simpler patterns appear to have smaller $S$ which means they mostly go through the low-resolution branches in each block.

\begin{figure*}[t]
\begin{center}
   \includegraphics[width=0.95\linewidth]{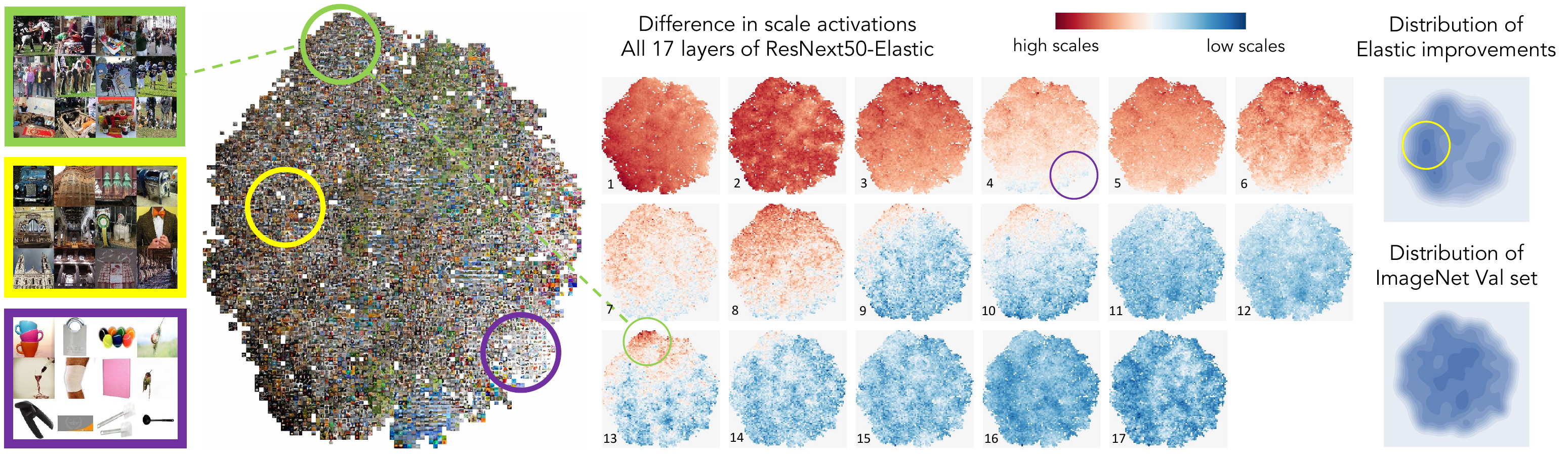}
\end{center}
   \vspace{-0.5cm}\caption{\small\textbf{Scale policy analysis.} This figure shows the impact of the scale policy on the accuracy of our Elastic model. (left) shows all the ImageNet validation set clustered using tsne by their scale policy pattern in the ResNeXt50+Elastic as discussed in section \ref{sec:policyanal}. (middle) shows the the scale policy score of all the images at 17 blocks of the network. Most of the images use high-resolution features at early layers and low-resolution features at later layers but some images break this pattern. Images pointed in the {\color{green}green circle} use high-resolution features in the $13^{th}$ block. Images pointed in the {\color{purple}purple circle} use low-resolution features in the $4^{th}$ block. These images usually contain a simpler pattern. (right)-bottom shows the density of images in the tsne space and (right)-top shows the density of the images that got correctly classified by Elastic model but miss-classified by the base ResNeXt model. This shows that Elastic can improve prediction when images are challenging in terms of their scale information. Some samples are pointed by the {\color{yellow}yellow circle}. Best viewed in color.}     
\label{fig:tsne}
\vspace{-0.2cm}
\end{figure*}

To analyze the impact of the scale policy on the accuracy of the Elastic, we represent each image (in the ImageNet validation set) by a 17-dimensional vector such that the values of the 17 elements are the scale policy score $S$ for the 17 Elastic blocks in a ResNeXt50+Elastic model. Then we apply tsne\cite{maaten2008visualizing} on all these vectors to get a two-dimensional visualization. In figure~\ref{fig:tsne}-(left) we draw all the images in the tsne coordinates. It can be seen that images are clustered based on their complexity pattern. In figure~\ref{fig:tsne}-(middle) for all of the images we show the 17 scale policy scores $S$ in 17 blocks. As it can be seen most of the images go through the high-resolution branches on the early layers and low-resolution branches at the later layers but some images break this pattern. For examples, images pointed by the green circle are activating high-resolution branches in the $13^{th}$ block of the network. These images usually contain a complex pattern that the network needs to extract features in high-resolution to classify correctly. Images pointed by the purple circle are activating low-resolution branches at early layers, the $4^{th}$ block of the network. These images usually contain a simple pattern that the network can classify at low-resolution early on. In Figure~\ref{fig:tsne}-(right) we show the density of all validation images in the tsne space in the bottom row, and in the top row, we show the density of images that are correctly classified by our Elastic model and miss-classified by the base ResNeXt model. This comparison shows that most of the images that Elastic can improve predictions on are the ones with more challenging scale properties. Some of them are pointed out by the yellow circle.
\subsection{MS COCO multi-label classification}
\begin{figure}[t]
\centering
    \begin{subfigure}[b]{0.23\textwidth}
        \includegraphics[width=\textwidth]{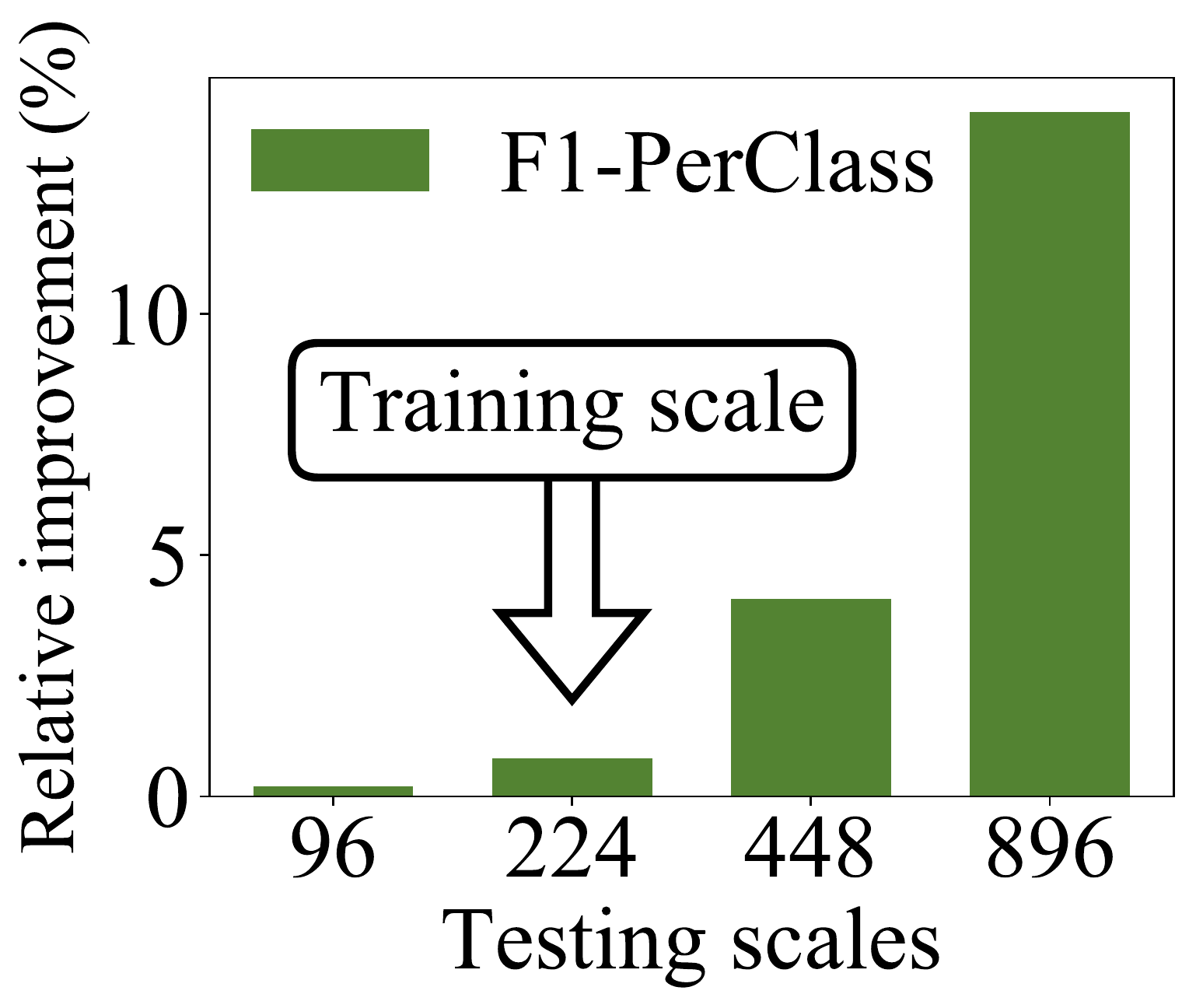}
        \label{fig:stressFC}
    \end{subfigure}~
    \begin{subfigure}[b]{0.22\textwidth}
        \includegraphics[width=\textwidth]{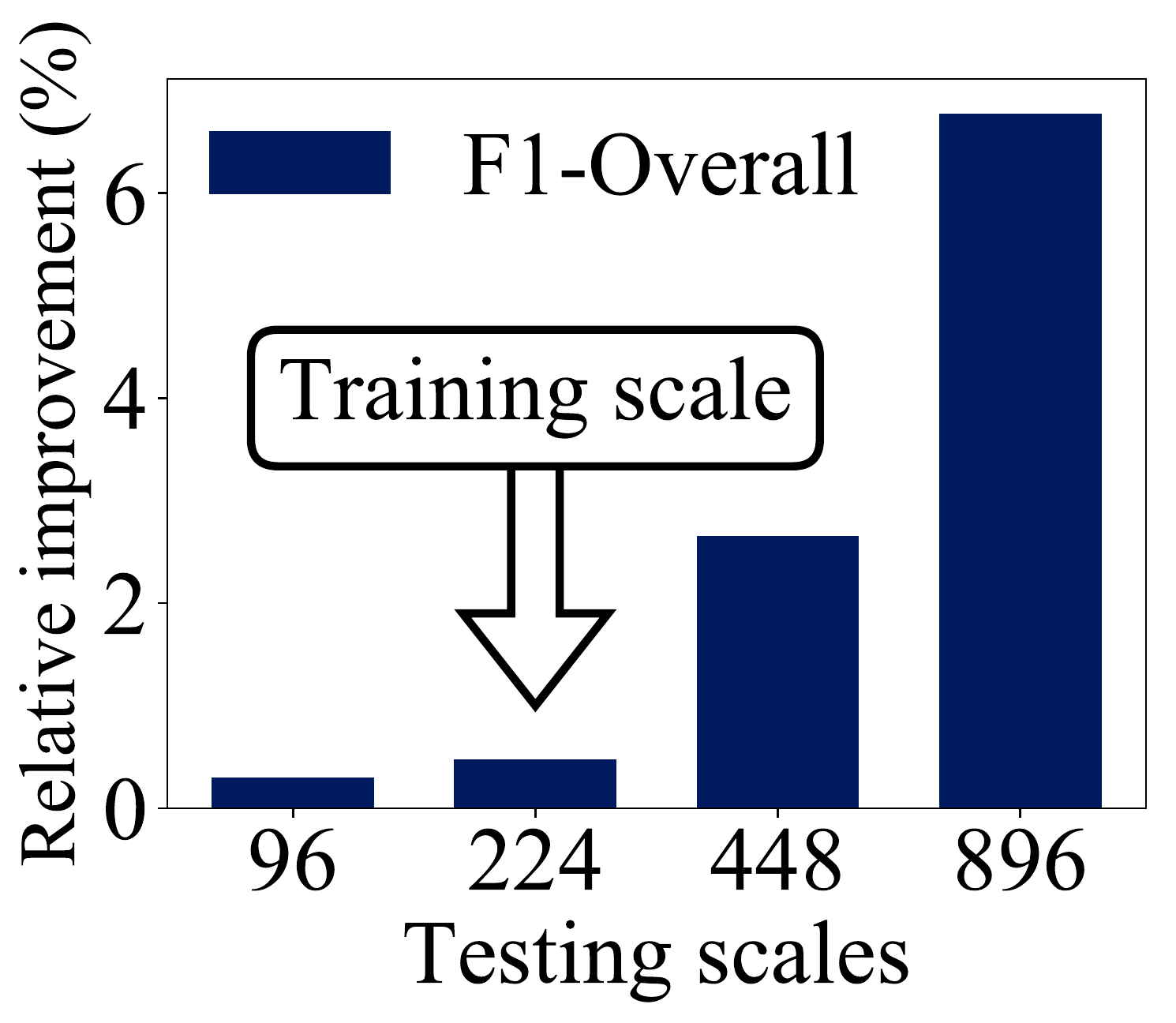}
        \label{fig:stressFO}
    \end{subfigure}
    \vspace{-0.6cm}
    \caption{\small\textbf{Scale stress test} on MSCOCO multi-label classification. This bar chart shows the relative F1 improvement of DLA-x60 being augmented Elastic over different image resolutions. Although both models are trained on $224\times224$ images, Elastic shows larger improvement when tested on high-resolution images.}
\label{fig:stress}
\vspace{-0.0cm}
\end{figure}

To further investigate the generality of our model, we finetune our ImageNet pre-trained model and evaluate on MS COCO multi-label classification task. The MSCOCO images are far more complicated in that there exist multiple objects from different categories and scales in each image.  
\vspace{-0.4cm}\paragraph{Implementation details.}
All models that we report are finetuned from ImageNet pre-trained model for 36 epochs with learning rate starting at 0.001 and being divided by 10 at epoch 24, 30. We train on 4 workers and 24 images per worker with SGD and weight decay of 0.0005. We train our models with binary cross entropy (BCE) loss, which is usually used as a baseline for domain-specific works that explicitly model spatial or semantic relations. We use the same data augmentations as our ImageNet training, and adopt standard multi-label testing on images resized to $224\times224$.
\vspace{-0.4cm}\paragraph{Evaluation metrics.}
Following the literature of multi-label classification\cite{zhu2017learning,ge2018multi,zhang2018multi,li2017improving}, results are evaluated using macro/micro evaluations. After training the models with BCE loss, labels with greater than 0.5 probability are considered positive. Then, macro and micro F1-scores are calculated to measure overall performance and the average of per-class performances respectively.
\vspace{-0.4cm}\paragraph{Results.} Table ~\ref{tab:mscocof1}  shows that elastic consistently improves per-class F1 and overall F1. In the case of DLA, Elastic augmentation even reduces the FLOPs and number of parameters by a large margin.
\begin{table}
\centering
\resizebox{0.8\columnwidth}{!}{
\begin{tabular}{l|cc}
\toprule
Model & F1-PerClass & F1-Overall  \\
\midrule
ResNet101\textsuperscript{*} & 69.98 & 74.58 \\
\midrule
DenseNet201\textsuperscript{*}  & 69.95 & 74.50 \\
DenseNet201+Elastic & \textbf{70.40} & \textbf{74.99} \\
\midrule
DLA-X60\textsuperscript{*} & 70.79 & 75.41 \\
DLA-X60+Elastic & \textbf{71.35} & \textbf{75.77}\\
\midrule
ResNeXt50\textsuperscript{*} & 70.12 & 74.52 \\
ResNeXt50+Elastic & \textbf{71.08} & \textbf{75.37} \\
\midrule
ResNeXt101\textsuperscript{*} & 70.95 & 75.21 \\
ResNeXt101+Elastic & \textbf{71.83} & \textbf{75.93} \\
\bottomrule
\end{tabular}
}
\vspace{-0.2cm}
\caption{\small\textbf{MSCOCO multi-class classification.} This table shows the generality of our Elastic model by finetuning pre-trained ImageNet models on MSCOCO multi-class images with binary cross entropy loss. Elastic improves F1 scores all across the board. }
\label{tab:mscocof1}
\vspace{-0.2cm}
\end{table}
\vspace{-0.4cm}\paragraph{Scale challenging images.}
We claimed that Elastic is very effective on scale challenging images. Now, we empirically show that a large portion of the accuracy improvement of our Elastic model is rooted in a better scale policy learning. We follow MSCOCO official split of \textit{small}, \textit{medium}, and \textit{large} objects. Per-class and overall F1, on small, medium and large objects, are computed. Since we don't have per-scale predictions, false positives are shared and re-defined as cases where none of small, medium, large object appears, but the model predicts positive. Results in Table \ref{tab:sml} show that ResNeXt50 + Elastic provides the largest gains on small objects. Elastic allows large objects to be dynamically captured by low resolution paths, so filters in high resolution branches do not waste capacity dealing with parts of large objects. Elastic blocks also merge various scales and feed scale-invariant features into the next block, so it shares computation in all higher blocks, and thus allows more capacity for small objects, at high resolution. This proves our hypothesis that Elastic understands scale challenging images better through scale policy learning.
\begin{table}[b]
\centering
\vspace{-0.5cm}
\resizebox{1.0\columnwidth}{!}{
\begin{tabular}{c|ccc|ccc}
\toprule
Model & Sm-C & Md-C & Lg-C & Sm-O & Md-O & Lg-O \\
\midrule
ResNeXt50 & 45.57 & 61.99 & 65.88 & 58.51 & 68.51 & 77.53\\
+Elastic & 46.67 & 63.05 & 66.46 & 59.47 & 69.47 & 78.03\\
\midrule
Relative & 2.43\% & 1.72\% & 0.88\% & 1.64\% & 1.40\% & 0.65\% \\
\bottomrule
\end{tabular}
}
\vspace{-0.1cm}
\caption{\small F1 scores on small, medium, and large objects respectively. C means per-class F1 and O means overall F1. ResNeXt50 + Elastic improves the most on small objects.}
\label{tab:sml}
\end{table}
\vspace{-0.4cm} \paragraph{Scale stress test.}
Besides standard testing where images are resized to $224\times224$, we also perform a stress test on the validation set. MSCOCO images' resolutions are \textapprox $640\times480$. Given a DLA-X60 model trained with $224\times224$ images, we also test it with images from different resolutions: $96\times96$, $448\times448$, $896\times896$ and change the last average pooling layer accordingly. Figure~\ref{fig:stress} shows that Elastic does not only perform well on trained scale, but also shows greater improvement on higher resolution images at test time. In addition, we do not observe an accuracy drop on $96\times96$ test, though the total computation assigned to low level is reduced in DLA-X60+Elastic.

\subsection{PASCAL VOC semantic segmentation}
To show the strength of our Elastic model on a pixel level classification task, we report experiments on PASCAL VOC semantic segmentation. ResNeXt models use weight decay 5e-4 instead of 1e-4 in ResNet. All models are trained for 50 epochs and we report mean intersection-over-union (IOU) on the val set. Other implementation details follow \cite{chen2017rethinking}, with MG(1, 2, 4), ASPP(6, 12, 18), image pooling, OS=16, batch size of 16, for both training and validation, without bells and whistles. Our ResNet101 reproduces the mIOU of 77.21\% reported in \cite{chen2017rethinking}. Our DLA models use the original iterative deep aggregation as a decoder and are trained with the same scheduling as \cite{chen2017rethinking}. In Table. \ref{tab:segmentation}, Elastic shows a large margin of improvement. This verifies that Elastic finds the scale policy that allows processing high-level semantic information and low-level boundary information together, which is critical in the task of semantic segmentation.
\begin{table}
\centering
\resizebox{0.6\columnwidth}{!}{
\begin{tabular}{l|cc}
\toprule
Model & Original & Elastic \\
\midrule
ResNeXt50\textsuperscript{*} & 75.29 & \textbf{77.70} \\
ResNeXt101\textsuperscript{*} & 77.47 & \textbf{78.51}\\
DLA-X60\textsuperscript{*} & 69.96 & \textbf{73.59} \\
\bottomrule
\end{tabular}
}
\vspace{-0.2cm}
\caption{\small\textbf{PASCAL VOC semantic segmentation}. This table compares the accuracy of semantic image segmentation (mIOU\%) using Elastic models vs. the original model. Elastic models outperform original models by a large margin. This supports that Elastic learns a scale policy that allows processing high-level semantic information and low-level boundary information together.}
\label{tab:segmentation}
\vspace{-0.2cm}
\end{table}
\subsection{Ablation study}
\label{sec:ablation}
In this section, we study the effect of different elements in Elastic models. We chose DLA-X60 as our baseline and applied Elastic to perform the ablation experiments.    
\vspace{-0.4cm}\paragraph{Upsampling/Downsampling methods.} We carried our experiments with bilinear up(down)sampling on DLA-X60+Elastic. In Table \ref{tab:ablation1} we show the accuracy of ImageNet classification using Elastic by different choices of up(down)sampling methods: Bilinear, Nearest, Trained filters and Trained Dilated filters with and without average pooling (indicated by \textbf{w/ AP}). Our experiment shows Elastic with the bilinear up(down)sampling is the best choice. 
\begin{table}
\centering
\resizebox{0.9\columnwidth}{!}{
\begin{tabular}{l|cc}
\toprule
Method & \# FLOPs & Top-1 error \\
\midrule
Original (no Elastic) & 3.6B & 21.92 \\
Bilinear w/ AP & \textbf{3.2B} & 21.25 \\
Nearest w/ AP & 3.2B & 21.49 \\
Trained Dilated Filter w/ AP & 3.6B & \textbf{21.20} \\
Trained Dilated Filter & 3.6B & 21.60 \\
Trained Filter & 3.2B & 21.52 \\
\bottomrule
\end{tabular}
}
\vspace{-0.2cm}
\caption{\small\textbf{Ablation study of up(down)sampling methods.} In this table, we show the accuracy of ImageNet classification using Elastic by different choices of up(down)sampling methods.  \textbf{w/ AP} indicates average pooling. Our experiment shows Elastic with bilinear up(down)sampling is the best choice with reduced FLOPs. }
\label{tab:ablation1}
\vspace{-0.1cm}
\end{table}
\vspace{-0.8cm}\paragraph{High/low-resolution branching rate.} We sweep over different choices of dividing parallel branches in the blocks into the high and low-resolutions.  In table \ref{tab:ablation2} we compare the variations of the percentage of branches allocated to high and low-resolutions at each block. This experiment shows that the best trade-off is when we equally divide the branches into high and low-resolutions. Interestingly, all of the branching options are outperforming the vanilla model (without Elastic). This shows that our Elastic model is quite robust to this parameter.   
\begin{table}
\centering
\resizebox{0.8\columnwidth}{!}{
\begin{tabular}{cc|cc}
\toprule
High-Res & Low-Res & FLOPs & Top-1 error \\
\midrule
100\% & 0\% & 3.6B & 21.92 \\
50\% & 50\% & 3.2B & 21.25 \\
75\% & 25\% & 3.4B & 21.35 \\
25\% & 75\% & 2.9B & 21.44 \\
\bottomrule
\end{tabular}
}
\vspace{-0.2cm}
\caption{\small\textbf{Ablation study of high(low) resolution branching rates}. In this table, we evaluate different branching rate across high and low-resolutions at each block. We observe that the best trade-off is when we equally divide the branches into high and low-resolutions. Independent of the ratio, all variations of branching are better than the base model.}
\label{tab:ablation2}
\vspace{-0.3cm}
\end{table}
\section{Conclusion}
\label{sec:conclusion}
We proposed Elastic, a model that captures scale variations in images by learning the scale policy from data. Our Elastic model is simple, efficient and very effective. Our model can easily be applied to any CNN architectures and improve accuracy while maintaining the same computation (or lower) as the original model. We applied Elastic to several state-of-the-art network architectures and showed consistent improvement on ImageNet classification, MSCOCO multi-class classification, and PASCAL VOC semantic segmentation. Our results show major improvement for images with scale challenges e.g. images consist of several small objects or objects with large scale variations.
\section*{Acknowledgments}
We thank Zhishuai Zhang, Siyuan Qiao, and Chenxi Liu for
their insightful comments and suggestions. This work is in part supported by ONR N00014-12-1-0883, NSF IIS-165205,  NSF IIS-1637479, NSF IIS-1703166, Sloan Fellowship, NVIDIA Artificial Intelligence Lab, and Allen Institute for artificial intelligence.

{\small
\bibliographystyle{ieee}
\bibliography{99_references}
}
\newpage
\onecolumn

\begin{appendices}
\section{sElastic (simple Elastic)}
A simple way of augmenting current models with Elastic is directly replacing bottlenecks by Elastic bottlenecks. This leads to models with less FLOPs and exactly the same number of parameters, which we refer to as sElastic (simple Elastic). This is in comparison to Elastic models that maintain the number of FLOPs and parameters. As shown in Table \ref{tab:ne}, sElastic already outperforms some of the original models, with less FLOPs. Note that DLA-X60+sElastic in Table \ref{tab:ne} is equivalent to DLA-X60+Elastic (in Table~\ref{tab:imagenet} in the original paper), i.e. we do not add/remove layers in different scales.\par

\begin{table}[h!]
\setlength\tabcolsep{3pt}
\begin{center}
\begin{tabular}{l|cc|cc}
\toprule
Model & \# Params & FLOPs & Top-1 & Top-5 \\
\midrule
ResNext50 & 25.0M & 4.2B & 22.2 & - \\
ResNext50\textsuperscript{*} & 25.0M & 4.2B & 22.23 & 6.25 \\
ResNext50+sElastic & 25.0M & 3.4B & \textbf{22.03} & \textbf{6.07} \\
\color{gray}ResNeXt50+Elastic & \color{gray}25.2M & \color{gray}4.2B & \color{gray}21.56 & \color{gray}5.83 \\
\midrule
DLA-X60 & 17.6M & 3.6B & 21.8 & - \\
DLA-X60\textsuperscript{*} & 17.6M & 3.6B & 21.92 & 6.03 \\
DLA-X60+sElastic & 17.6M & 3.2B & \textbf{21.25} & \textbf{5.71} \\
\color{gray}DLA-X60+Elastic & \color{gray}17.6M & \color{gray}3.2B & \color{gray}21.25 & \color{gray}5.71 \\
\midrule
DLA-X102 & 26.8M & 6.0B & 21.5 & - \\
DLA-X102+sElastic & 26.8M & 5.0B & \textbf{21.0} & \textbf{5.66} \\
\color{gray} DLA-X102+Elastic & \color{gray}24.9M & \color{gray}6.0B & \color{gray}20.71 & \color{gray}5.38 \\
\bottomrule
\end{tabular}
\end{center}
\vspace{-0.5cm}
\caption{\textbf{Error rates for sElastic on the ImageNet validation set.} sElastic models with reduced FLOPs already perform better than some of the original models. We also provide the Elastic versions from the original paper as a reference.}
\label{tab:ne}
\end{table}

\section{Elastic Architecture Details}
SElastic already outperforms original models. However, only applying downsamplings equivalently shifts computation from low level to higher level, which could cause lack of low level features to support high level processing. Also, sElastic reduces FLOPs so that its accuracy is not fairly comparable with the original model. For these two reasons, we rearrange computation distribution in each resolution, and this leads to our final Elastic model. \par
Consider ResNeXt-50 as an example. The original model assigns [3, 4, 6, 3] blocks respectively to [56, 28, 14, 7] four scales. As shown in Table \ref{tab:supp_resnext}, sElastic simply replaces original bottlenecks with Elastic bottlenecks. In Elastic, we roughly match the scale distribution of the original model by assigning [6, 8, 5, 3] blocks to those resolutions, as shown in Table \ref{tab:supp_resnext}. Note that half of each block processes information at a higher level. This modification also leads to matched number of parameters, and matched number of FLOPs. For ResNeXt101, we use a block design of [12, 14, 20, 3]. DenseNet+Elastic and DLA+Elastic architectures are shown respectively in Table \ref{tab:supp_densenet} and Table \ref{tab:supp_dla}. Note that these block designs were picked to match the original number of parameters and FLOPs, so we didn't tune them as hyper-parameters. Tuning them could probably lead to even lower error rates.

\begin{table*}[h!]
\begin{center}
\resizebox{0.95\textwidth}{!}{
\setlength\tabcolsep{5pt}
\begin{tabular}{l|l|cccccc|cc}
\toprule
Name & Block & Stage 1 & Stage 2 & Stage 3 & Stage 4 & Stage 5 & Stage 6 & Params. & FLOPs\\
\midrule
DLA-X60 & Split32 & 16 & 32 & 1-128 & 2-256 & 3-512 & 1-1024 & 17.7 $\times10^6$ & 3.6 $\times10^9$ \\
DLA-X60+Elastic & Split32+Elastic & 16 & 32 & 1-128 & 2-256 & 3-512 & 1-1024 & 17.7 $\times10^6$ & 3.2 $\times10^9$  \\
\midrule
DLA-X102 & Split32 & 16 & 32 & 1-128 & 3-256 & 4-512 & 1-1024 & 26.8 $\times10^6$ & 6.0 $\times10^9$  \\
DLA-X102+sElastic & Split32+Elastic & 16 & 32 & 1-128 & 3-256 & 4-512 & 1-1024 & 26.8 $\times10^6$ & 5.0 $\times10^9$  \\
DLA-X102+Elastic & Split50+Elastic & 16 & 32 & 3-128 & 3-256 & 3-512 & 1-1024 & 24.9 $\times10^6$ & 6.0 $\times10^9$  \\
\bottomrule
\end{tabular}
}
\end{center}
\caption{\textbf{DLA model architectures.} Following DLA, we show our  DLA classification architectures in the table. Split32 means a ResNeXt bottleneck with 32 paths while Split50 means a ResNeXt bottleneck with 50 paths. Stages 3 to 6 show d-n where d is the aggregation depth and n is the number of channels.}
\label{tab:supp_dla}
\end{table*}
\begin{table*}[h!]
\begin{center}
\setlength\tabcolsep{3pt}
\resizebox{0.95\textwidth}{!}{
\begin{tabular}{|c|c|c|c|}
\hline
stage & ResNeXt50 & ResNeXt50+sElastic & ResNeXt50+Elastic \\
\hline
conv1 & \multicolumn{3}{|c|}{7$\times$7, 64, stride 2, 3$\times$3 max pool, stride 2} \\
\hline
\begin{tabular}{ c }
  conv2 \\
  56$\times$56
\end{tabular}
&\begin{math}\begin{bmatrix}
    \begin{tabular}{c}
    1$\times$1, 128  \\
    3$\times$3, 128, C=32 \\
       1$\times$1, 256  
    \end{tabular}\end{bmatrix}\times\end{math} 3
&\begin{math}\begin{bmatrix}
    \begin{tabular}{ccc}
    & & 2$\times$down, 28$\times$28 \\
    1$\times$1, 64 & & 1$\times$1, 64           \\
    3$\times$3, 64, C=16 &+& 3$\times$3, 64, C=16\\
       1$\times$1, 256      & & 1$\times$1, 256     \\
    & & 2$\times$up, 56$\times$56
    \end{tabular}\end{bmatrix}\times\end{math} 3
&\begin{math}\begin{bmatrix}
    \begin{tabular}{ccc}
    & & 2$\times$down, 28$\times$28 \\
    1$\times$1, 64 & & 1$\times$1, 64           \\
    3$\times$3, 64, C=16 &+& 3$\times$3, 64, C=16\\
       1$\times$1, 256      & & 1$\times$1, 256     \\
    & & 2$\times$up, 56$\times$56
    \end{tabular}\end{bmatrix}\times\end{math} 6 \\
\hline
\begin{tabular}{ c }
  conv3 \\
  28$\times$28 \\
\end{tabular} 
&\begin{math}\begin{bmatrix}
    \begin{tabular}{c}
    1$\times$1, 256          \\
    3$\times$3, 256, C=32 \\
       1$\times$1, 512  
    \end{tabular}\end{bmatrix}\times\end{math} 4
& \begin{math}\begin{bmatrix}\begin{tabular}{ccc}
       & & 2$\times$ down, 14 $\times$ 14\\
       1$\times$1, 128       & & 1$\times$1, 128           \\
       3$\times$3, 128, C=16 &+& 3$\times$3, 128, C=16    \\
       1$\times$1, 512       & & 1$\times$1, 512     \\
       & & 2$\times$ up, 28 $\times$ 28
     \end{tabular}\end{bmatrix}\times\end{math} 4
& \begin{math}\begin{bmatrix}\begin{tabular}{ccc}
       & & 2$\times$ down, 14 $\times$ 14\\
       1$\times$1, 128       & & 1$\times$1, 128           \\
       3$\times$3, 128, C=16 &+& 3$\times$3, 128, C=16    \\
       1$\times$1, 512       & & 1$\times$1, 512     \\
       & & 2$\times$ up, 28 $\times$ 28
     \end{tabular}\end{bmatrix}\times\end{math} 8\\
\hline
\begin{tabular}{ c }
  conv4 \\
  14$\times$14 \\
\end{tabular} 
&\begin{math}\begin{bmatrix}
    \begin{tabular}{c}
    1$\times$1, 512    \\
    3$\times$3, 512, C=32\\
       1$\times$1, 1024  
    \end{tabular}\end{bmatrix}\times\end{math} 6
& \begin{math}\begin{bmatrix}\begin{tabular}{ccc}
       & & 2$\times$ down, 7 $\times$ 7\\
       1$\times$1, 256       & & 1$\times$1, 256           \\
       3$\times$3, 256, C=16 &+& 3$\times$3, 256, C=16    \\
       1$\times$1, 1024       & & 1$\times$1, 1024     \\
       & & 2$\times$ up, 14 $\times$ 14
     \end{tabular}\end{bmatrix}\times\end{math} 6
& \begin{math}\begin{bmatrix}\begin{tabular}{ccc}
       & & 2$\times$ down, 7 $\times$ 7\\
       1$\times$1, 256       & & 1$\times$1, 256           \\
       3$\times$3, 256, C=16 &+& 3$\times$3, 256, C=16    \\
       1$\times$1, 1024       & & 1$\times$1, 1024     \\
       & & 2$\times$ up, 14 $\times$ 14
     \end{tabular}\end{bmatrix}\times\end{math} 5\\
\hline
\begin{tabular}{ c }
  conv5 \\
  7$\times$7 \\
\end{tabular} &
\multicolumn{3}{|c|}{
\begin{math}\begin{bmatrix}\begin{tabular}{c}
       1$\times$1, 1024 \\
       3$\times$3, 1024, C=32 \\
       1$\times$1, 2048
     \end{tabular}\end{bmatrix} \times\end{math} 3}\\
\hline
1$\times$1 & \multicolumn{3}{|c|}{global average pool, 1000-d fc, softmax}\\
\hline
Params. & 25.0 $\times10^6$& 25.0 $\times10^6$& 25.2 $\times10^6$\\
\hline
FLOPs & 4.2 $\times10^9$& 3.4 $\times10^9$& 4.2 $\times10^9$\\
\hline
\end{tabular}
}
\end{center}
\caption{\textbf{ResNeXt50 vs. ResNeXt50+sElastic vs. ResNeXt50+Elastic.} ResNeXt50+Elastic employs two resolutions in each block, and keeps output resolution high for more blocks, compared with ResNeXt50.}
\label{tab:supp_resnext}
\end{table*}
\begin{table*}[h!]
\begin{center}
\setlength\tabcolsep{3pt}
\begin{tabular}{|c|c|c|}
\hline
stage & DenseNet201 & DenseNet201+Elastic \\
\hline
conv1 & \multicolumn{2}{|c|}{7$\times$7, 64, stride 2, 3$\times$3 max pool, stride 2} \\
\hline
\begin{tabular}{ c }
  conv2 \\
  56$\times$56
\end{tabular}
&\begin{math}\begin{bmatrix}
    \begin{tabular}{c}
    1$\times$1, 128  \\
    3$\times$3, 32 \\
    \end{tabular}\end{bmatrix}\times\end{math} 6
&\begin{math}\begin{bmatrix}
    \begin{tabular}{c}
    1$\times$1, 64 \\
    3$\times$3, 32
    \end{tabular}+\begin{tabular}{c}
    2$\times$down, 28$\times$28 \\
    1$\times$1, 64\\
    3$\times$3, 32\\
    2$\times$up, 56$\times$56
    \end{tabular}\end{bmatrix}\times\end{math} 10 \\
\hline
trans1 & \multicolumn{2}{|c|}{1$\times$1 conv, 2$\times$2 average pool, stride 2} \\
\hline
\begin{tabular}{ c }
  conv3 \\
  28$\times$28
\end{tabular}
&\begin{math}\begin{bmatrix}
    \begin{tabular}{c}
    1$\times$1, 128  \\
    3$\times$3, 32 \\
    \end{tabular}\end{bmatrix}\times\end{math} 12
&\begin{math}\begin{bmatrix}
    \begin{tabular}{c}
    1$\times$1, 64 \\
    3$\times$3, 32
    \end{tabular}+\begin{tabular}{c}
    2$\times$down, 14$\times$14 \\
    1$\times$1, 64\\
    3$\times$3, 32\\
    2$\times$up, 28$\times$28
    \end{tabular}\end{bmatrix}\times\end{math} 20 \\
\hline
trans2 & \multicolumn{2}{|c|}{1$\times$1 conv, 2$\times$2 average pool, stride 2} \\
\hline
\begin{tabular}{ c }
  conv4 \\
  14$\times$14
\end{tabular}
&\begin{math}\begin{bmatrix}
    \begin{tabular}{c}
    1$\times$1, 128  \\
    3$\times$3, 32 \\
    \end{tabular}\end{bmatrix}\times\end{math} 48
&\begin{math}\begin{bmatrix}
    \begin{tabular}{c}
    1$\times$1, 64 \\
    3$\times$3, 32
    \end{tabular}+\begin{tabular}{c}
    2$\times$down, 7$\times$7 \\
    1$\times$1, 64\\
    3$\times$3, 32\\
    2$\times$up, 14$\times$14
    \end{tabular}\end{bmatrix}\times\end{math} 40 \\
\hline
trans3 & \multicolumn{2}{|c|}{1$\times$1 conv, 2$\times$2 average pool, stride 2} \\
\hline
\begin{tabular}{ c }
  conv5 \\
  7$\times$7
\end{tabular}
&\begin{math}\begin{bmatrix}
    \begin{tabular}{c}
    1$\times$1, 128  \\
    3$\times$3, 32 \\
    \end{tabular}\end{bmatrix}\times\end{math} 32
&\begin{math}\begin{bmatrix}
    \begin{tabular}{c}
    1$\times$1, 128  \\
    3$\times$3, 32 \\
    \end{tabular}\end{bmatrix}\times\end{math} 30 \\
\hline
1$\times$1 & \multicolumn{2}{|c|}{global average pool, 1000-d fc, softmax}\\
\hline
Params. & 20.0 $\times10^6$& 19.5 $\times10^6$\\
\hline
FLOPs & 4.4 $\times10^9$& 4.2 $\times10^9$\\
\hline
\end{tabular}
\end{center}
\caption{\textbf{DenseNet201 vs. DenseNet201+Elastic.} DenseNet+Elastic follows a similar modification as ResNeXt+Elastic, i.e. two resolutions in each block and more blocks in high resolutions.}
\label{tab:supp_densenet}
\end{table*}

\section{Scale policy demo}
\begin{figure*}[h!]
\begin{center}
   \includegraphics[width=0.95\linewidth]{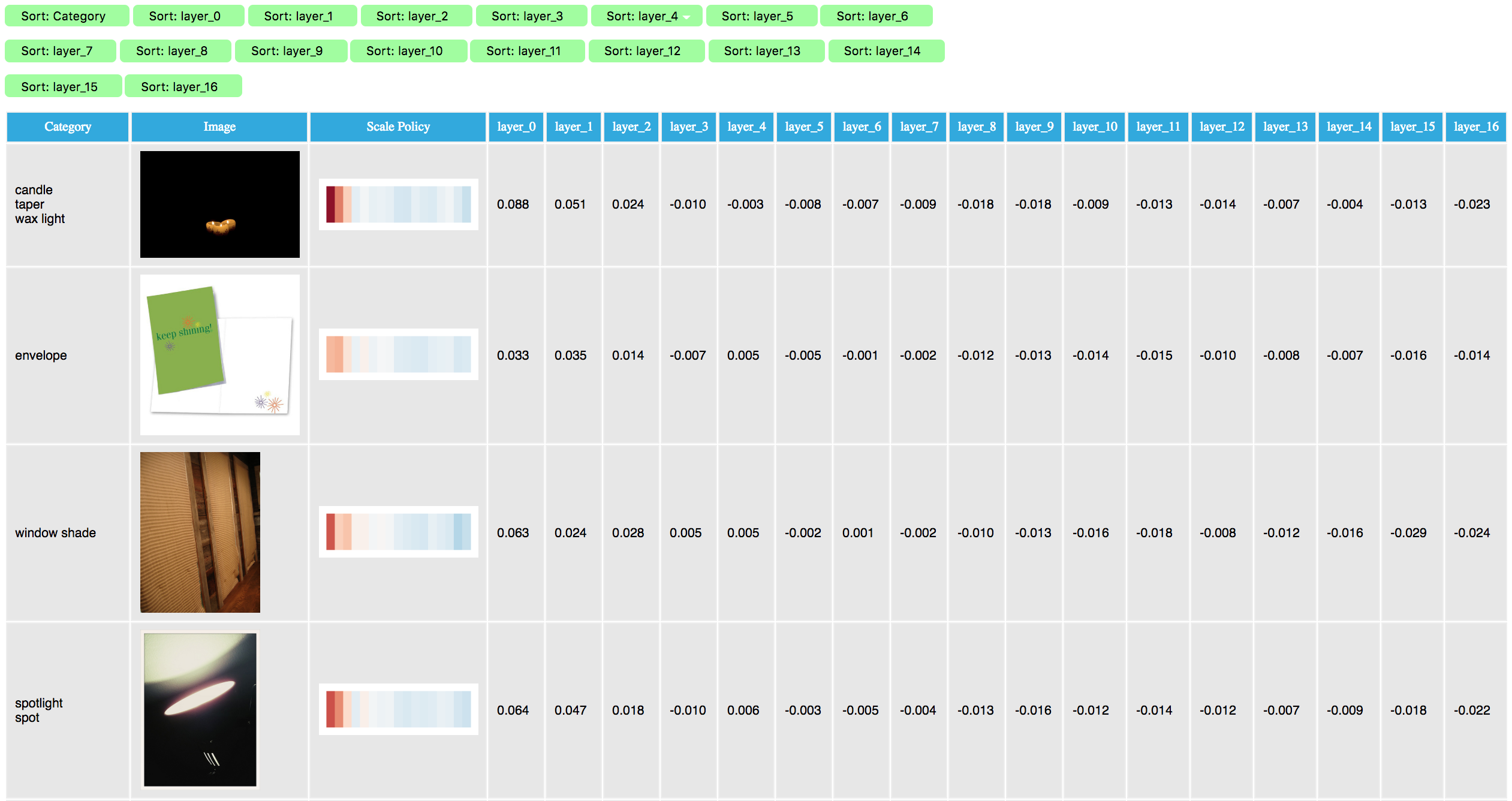}
\end{center}
\caption{\textbf{Screenshots of the scale policy demo.} Examples of low scale scores at layer 4. These images usually contain a simple pattern.}
\label{fig:demo1}
\end{figure*}

\begin{figure*}[h!]
\begin{center}
   \includegraphics[width=0.95\linewidth]{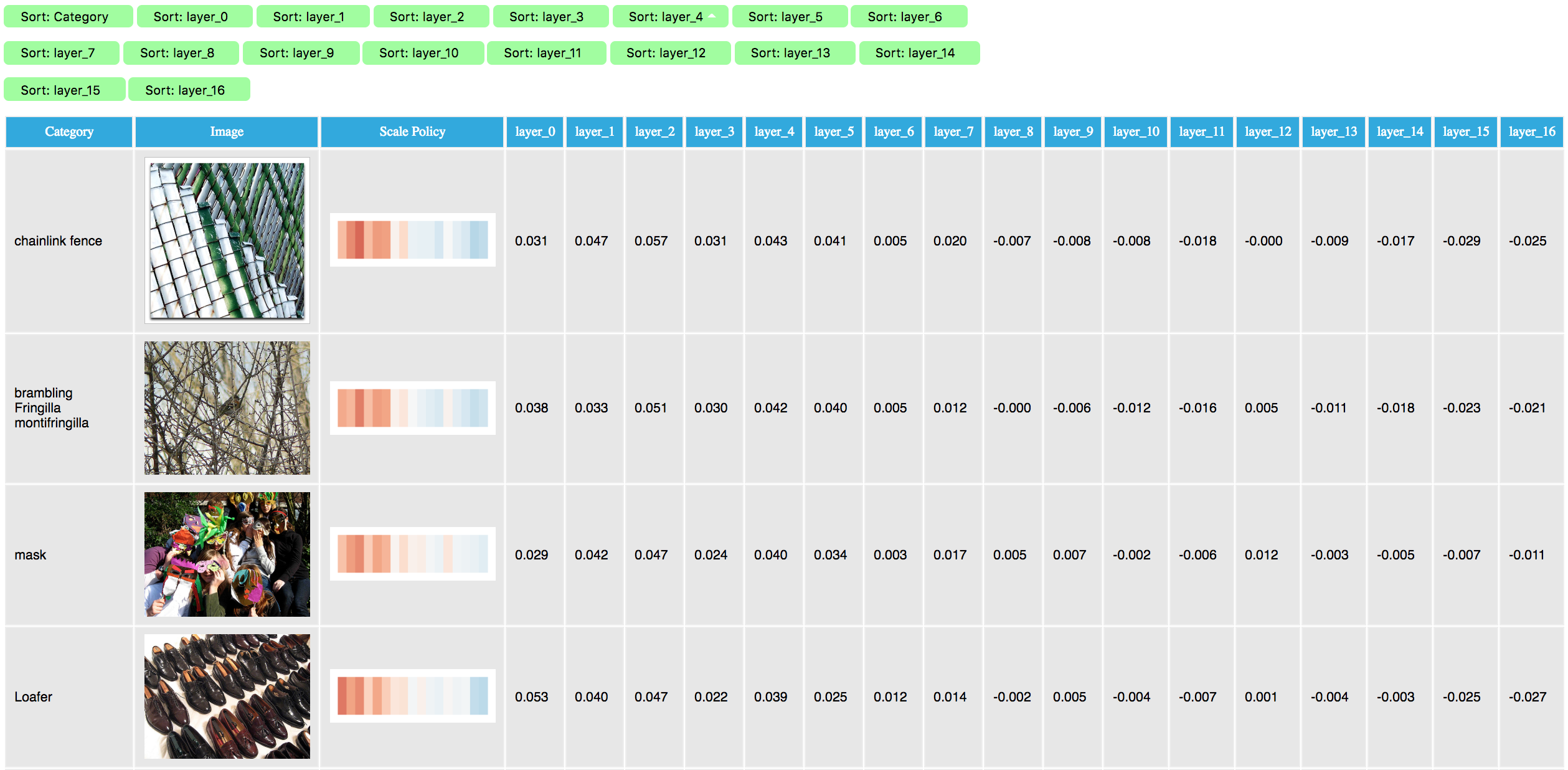}
\end{center}
\caption{\textbf{Screenshots of the scale policy demo.} Examples of high scale scores at layer 4. These images require detailed processing at high resolutions.}
\label{fig:demo2}
\end{figure*}

Apart from Figure~\ref{fig:teaser} and Figure~\ref{fig:tsne} in the main paper, we made an interactive HTML based demo of our learned scale policy, that allows a user to explore images in the validation set and their scale policies.
In the following screenshots we show some images where ResNeXt50+Elastic improves over the original ResNeXt50 on ImageNet validation set. Figures \ref{fig:demo1} and \ref{fig:demo2} show two screenshots. Each screenshot shows images with their classes, their scale policy visualizations, and their scale policy scores at all layers. The user can search through images and sort these images by their categories or their scale policy score at any layer. We refer interested reader to section \ref{sec:policyanal} of the main paper for the definition of scale policy score and more discussions on different scale policies. 

\section{Semantic segmentation results}
\begin{figure*}
    \centering
    
    \begin{subfigure}[b]{0.23\textwidth}
    \centering
    \setlength\tabcolsep{0.12345pt}
    \begin{tabular}{c}
        \includegraphics[width=0.95\textwidth]{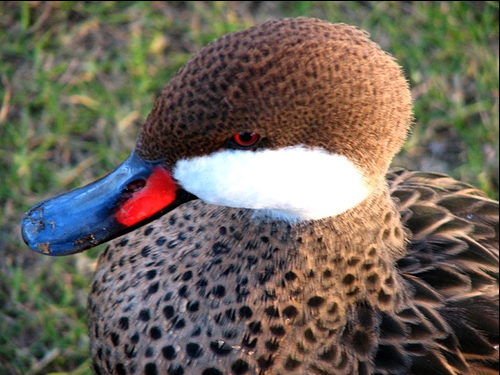}\\
        \includegraphics[width=0.95\textwidth]{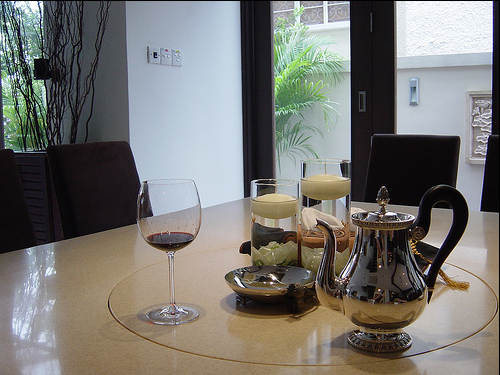}\\
        \includegraphics[width=0.95\textwidth]{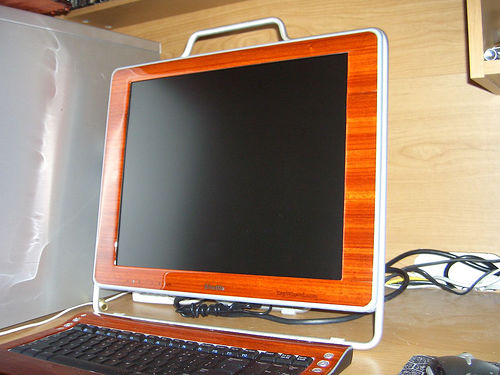}\\
        \includegraphics[width=0.95\textwidth]{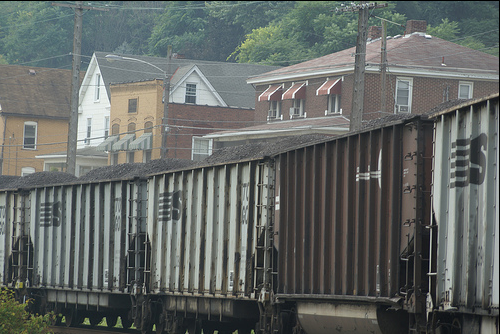}\\
        \includegraphics[width=0.95\textwidth]{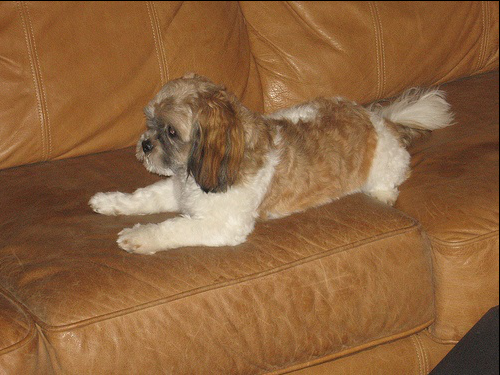}
    \end{tabular}
    \caption{Validation Images}
    \end{subfigure}
    \vrule
    \begin{subfigure}[b]{0.23\textwidth}
    \centering
    \setlength\tabcolsep{0.12345pt}
    \begin{tabular}{c}
        \includegraphics[width=0.95\textwidth]{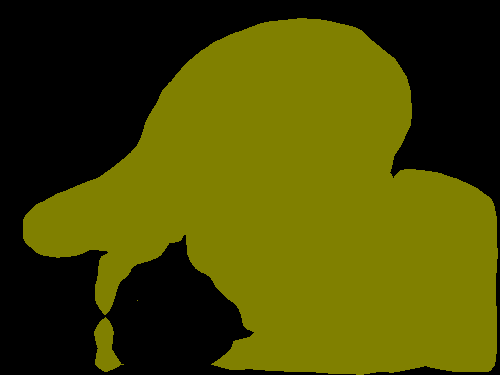}\\
        \includegraphics[width=0.95\textwidth]{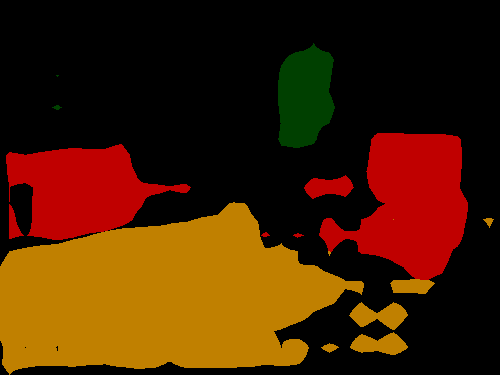}\\
        \includegraphics[width=0.95\textwidth]{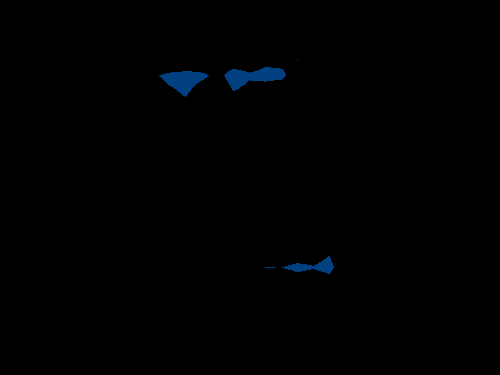}\\
        \includegraphics[width=0.95\textwidth]{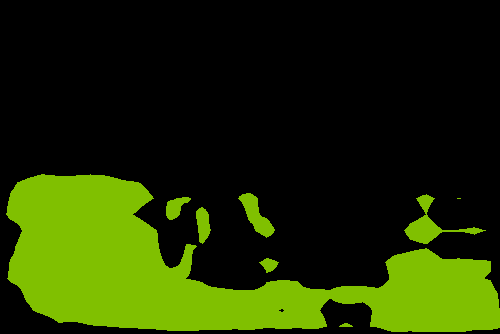}\\
        \includegraphics[width=0.95\textwidth]{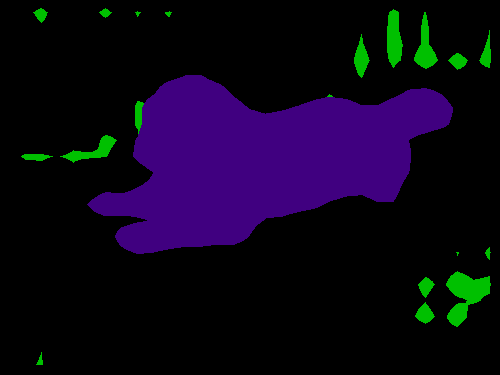}
    \end{tabular}
    \caption{ResNeXt101}
    \end{subfigure}
    \vrule
    \begin{subfigure}[b]{0.23\textwidth}
    \centering
    \setlength\tabcolsep{0.12345pt}
    \begin{tabular}{c}
        \includegraphics[width=0.95\textwidth]{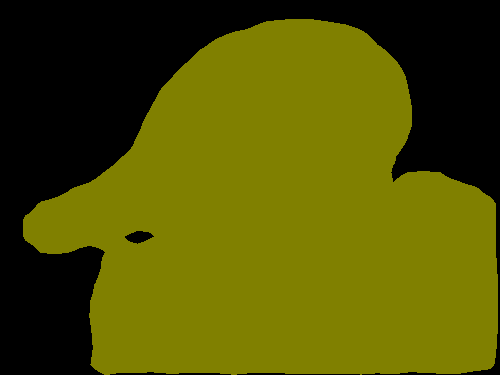}\\
        \includegraphics[width=0.95\textwidth]{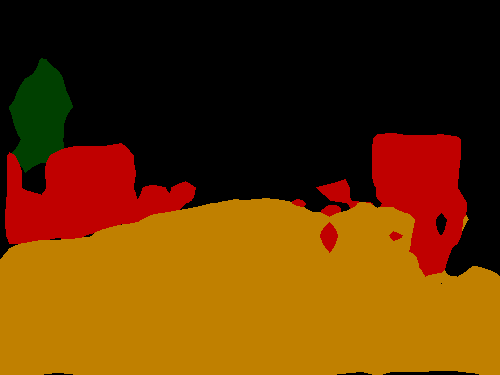}\\
        \includegraphics[width=0.95\textwidth]{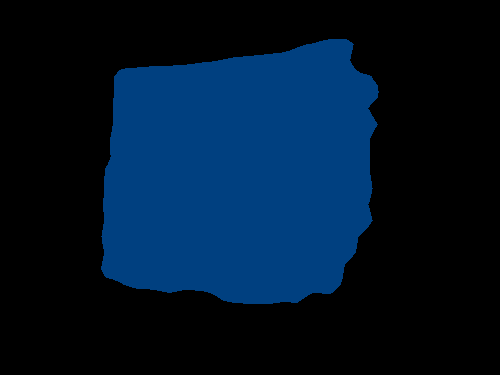}\\
        \includegraphics[width=0.95\textwidth]{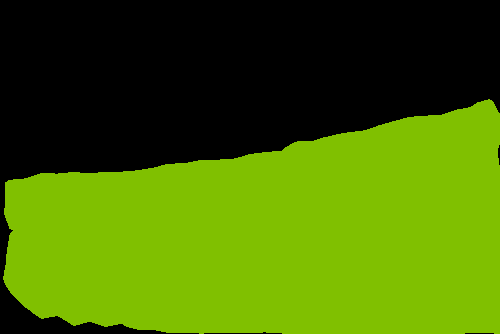}\\
        \includegraphics[width=0.95\textwidth]{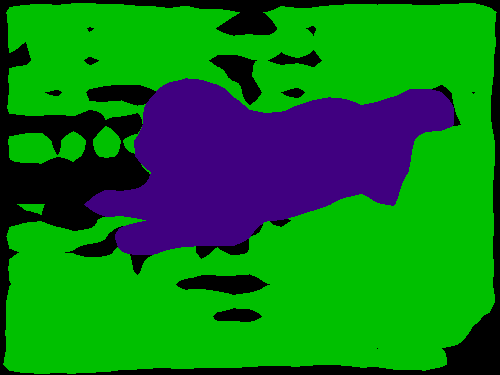}
    \end{tabular}
    \caption{ResNeXt101+Elastic}
    \end{subfigure}
    \vrule
    \begin{subfigure}[b]{0.23\textwidth}
    \centering
    \setlength\tabcolsep{0.12345pt}
    \begin{tabular}{c}
        \includegraphics[width=0.95\textwidth]{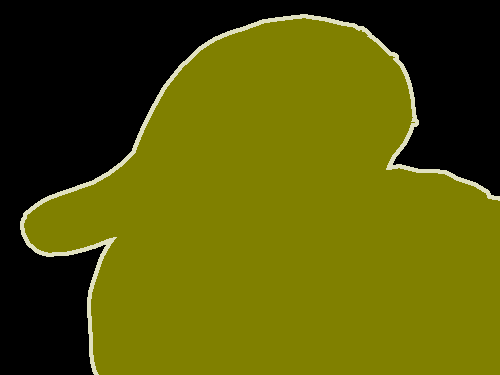}\\
        \includegraphics[width=0.95\textwidth]{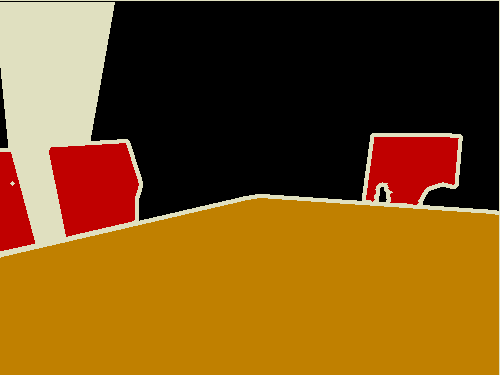}\\
        \includegraphics[width=0.95\textwidth]{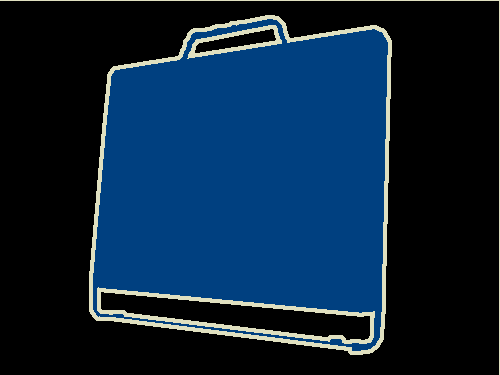}\\
        \includegraphics[width=0.95\textwidth]{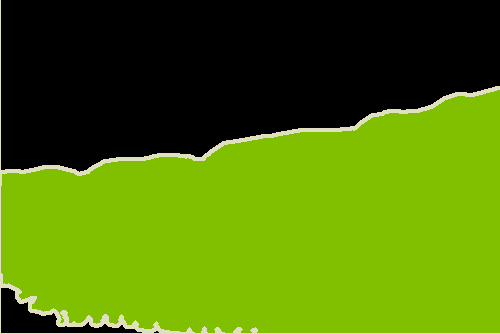}\\
        \includegraphics[width=0.95\textwidth]{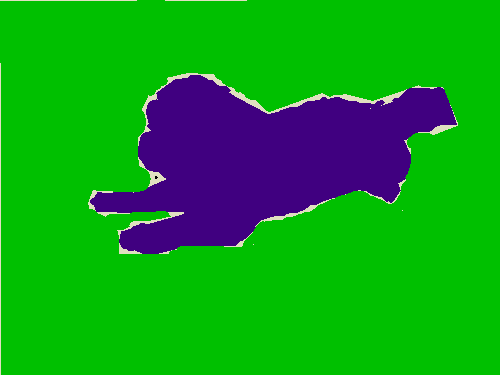}
    \end{tabular}
    \caption{Ground truth}
    \end{subfigure}
    
    \caption{\textbf{Semantic segmentation results on PASCAL VOC.} Elastic improves most on scale-challenging images.}
    \label{fig:segmentation}
\end{figure*}

Some visualizations of our semantic segmentation results are shown in Figure \ref{fig:segmentation}, demonstrating that Elastic segments scale-challenging objects well on PASCAL VOC.

\end{appendices}

\end{document}